\documentclass{article}

    \usepackage[preprint, nonatbib]{neurips_2020}

\usepackage[utf8]{inputenc} 
\usepackage[T1]{fontenc}    
\usepackage{hyperref}       
\usepackage{url}            
\usepackage{booktabs}       
\usepackage{amsfonts}       
\usepackage{nicefrac}       
\usepackage{microtype}      

\usepackage{graphicx}
\usepackage{subcaption}
\usepackage{algorithm}
\usepackage[noend]{algorithmic}
\usepackage{wrapfig}
\usepackage{tabularx}
\usepackage{mathtools}
\usepackage{amsmath}

\title{Interpreting Deep Temporal Neural Networks by Selective Visualization of Internally Activated Nodes}

\author{%
  Sohee Cho\thanks{Equal Contribution}, Wonjoon Chang, Jaesik Choi\thanks{Corresponding Author}\\
  Graduate School of AI\\
  KAIST \\
  Daejeon, Republic of Korea\\
  \{sohee.cho, one\_jj, jaesik.choi\}@kaist.ac.kr \\
  \And
  Ginkyeng Lee\footnotemark[1] \\
  Graduate School of ECE\\
  UNIST \\
  Ulsan, Republic of Korea\\
  gin908@unist.ac.kr \\
}

\begin{document}
\maketitle

\begin{abstract}
Recently deep neural networks have demonstrated competitive performance in classification and regression tasks for many types of sequential data. However, it is still hard to understand which temporal patterns the internal channels of deep neural networks see in sequential data. In this paper, we propose a new framework to visualize temporal representations learned in deep neural networks without hand-crafted segmentation labels. Given input data, our framework extracts highly activated temporal regions that contribute to activating internal nodes. Furthermore, our framework characterizes such regions by clustering and shows the representative temporal pattern for each cluster group with the uncertainty of data belonging to corresponding group. Thus, it enables users to identify whether the input has been observed frequently in the training data. In experiments, we verify that preserving the temporal regions selected by our framework makes internal nodes more robust than preserving the points selected by relevance scores in perturbation analysis.
\end{abstract}

\section{Introduction}
\hspace{3mm} The amount of temporal data has greatly increased due to the use of efficient and diverse automatic-information systems such as manufacturing sensors, stock database systems and healthcare wearable devices. Recent deep learning models enable users to extract appropriate features from such data and make decisions with high accuracy. As a result, to utilize such vast amounts of temporal data, demand for the application of deep learning models in the industry has grown rapidly. However, most industrial fields require transparency in decision-making when using deep learning models. Thus, they are still hesitant to adopt AI systems due to the lack of interpretability in their internal processes.

\hspace{3mm} Nowadays many AI researchers have tried to understand the decision process of deep learning models. Interpretable artificial intelligence methods explain and interpret decisions of complex systems with illustrative or textual descriptions. These approaches for deep learning are classified into explaining input attribution methods based on relevance score
\cite{binder2016layer, MONTAVON2017211, li2018patternnet, nam2019relative} , gradient-based methods \cite{shrikumar2017learning,springenberg2014striving,selvaraju2017grad} , explaining internal nodes methods \cite{bau2017network,bau2018gan} , explaining through attention methods \cite{choi2016retain, heo2018uncertainty} and generating explanations methods \cite{andreas2016neural}.

\hspace{3mm} Those methods mainly focus on the image domain, because visualized outputs with their own inputs enable users to intuitively interpret what the outputs mean. In Network Dissection ~\cite{bau2017network}, for example, input images are highlighted to show the spatial locations that each unit in the convolutional neural network is looking at. Layer-wise Relevance Propagation (LRP) ~\cite{MONTAVON2017211} explains the model's decision by decomposing an output of the model into individual pixels according to the amount of contribution to the output, so that people can visually understand complex concepts such as shapes or objects. However, there have been few efforts to apply interpretation techniques to time series data. Objects in images can be easily recognized visually, since there is a lot of human-annotated segmentation information for image datasets, which are not provided for most time series datasets. Thus, it is hard to find semi-global shapes that a neural network is looking at in a time series input due to a lack of temporally segmented annotation data.   

\hspace{3mm} To address this issue, we suggest a new framework to visualize temporal representations by clustering temporal patterns of highly activated nodes. Our framework has the following contributions:

\begin{itemize}
\item Without hand-crafted segmentation labels, our framework identifies representative temporal patterns that activate each channel of convolutional neural networks most.
\item Our framework matches perceived sub-sequences and the closest representative temporal patterns; It makes users easy to verify whether the patterns are commonly observed or deviated from trained data. 
\item Our framework provides the uncertainty of the representative temporal patterns. This uncertainty implies the potential variance of input shapes that activate the channel.
\end{itemize}

\hspace{3mm} Consequently, our framework can provide insight into the decision-making process of internal channels in the neural network through visualization. This visualization shows general shapes that the network recognizes with uncertainty. To the best of our knowledge, this is first attempt to extract and visualize patterns that highly activate interval nodes. It helps users to understand how deep neural networks learn time series data intelligibly. 

\begin{figure}[h]
\centering
    \begin{subfigure}[b]{\textwidth}\includegraphics[width =1\textwidth]{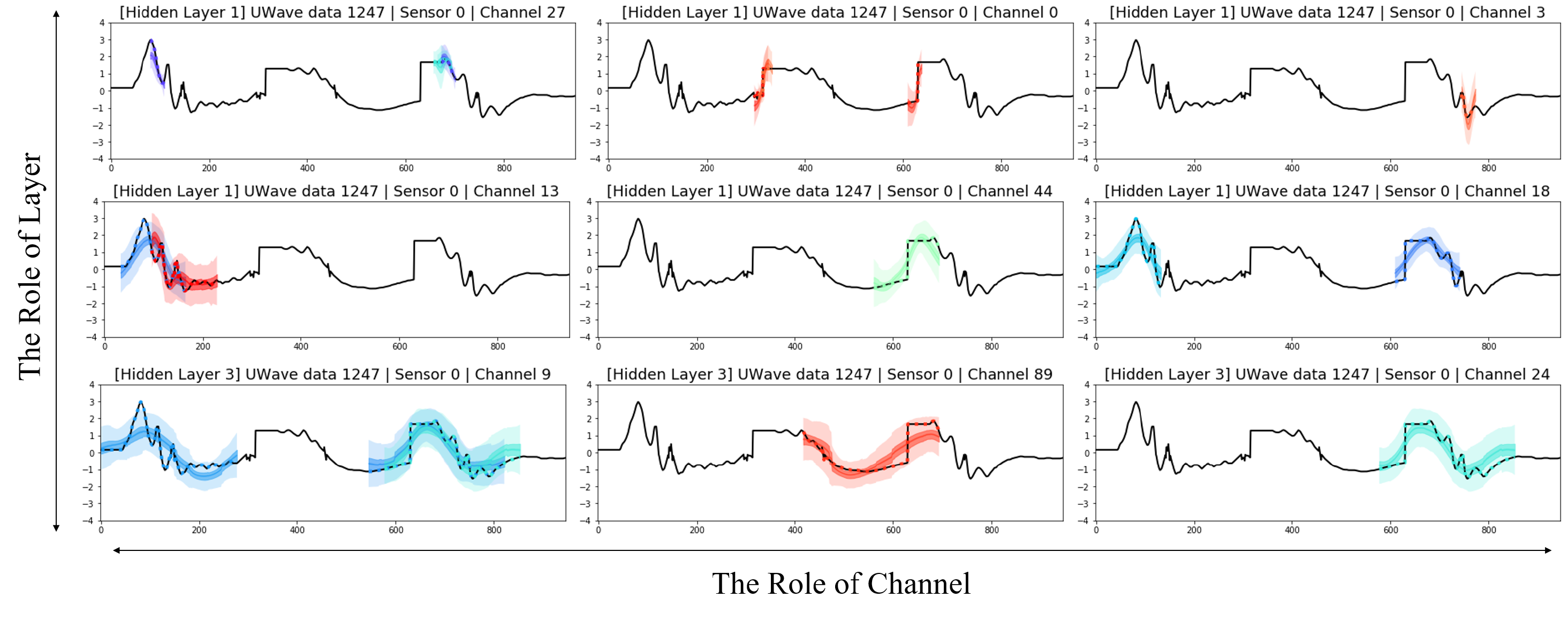} 
    \end{subfigure}
    \begin{subfigure}[b]{\textwidth}\includegraphics[width =1\textwidth]{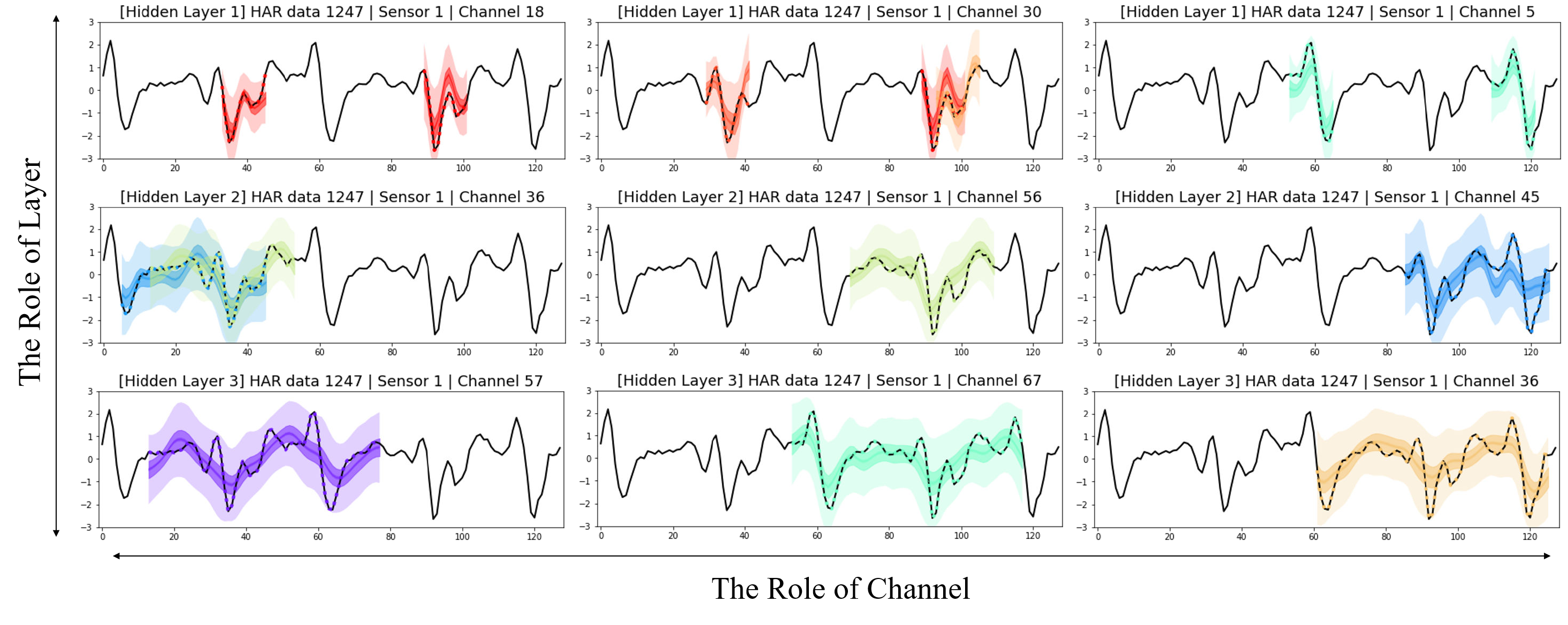} \end{subfigure}
    \caption{Our framework visualizes temporal representations learned from temporal deep neural networks. Given time series data, x-axis explains the role of channels that each channel of hidden layers extracts different features with other channels. Y-axis shows the role of hidden layer that high-level layers are affected by the long-term trend, while low-level layers are effected from detailed features.}
     \vspace{-5mm}
\end{figure}

\section{Related Work}

\hspace{3mm} In this section, This section briefly explains related work. Layer-wise Relevance Propagation (LRP) defines the output of the network as a relevance score, and then continues to compute the relevance score of previous layers recursively \cite{MONTAVON2017211}. Relevance scores of inputs become gradients multiplied with inputs when ReLU is used as the non-linear function. The idea is recursively applied to the previous layers from the output to decompose the relevance score into the sum of relevance scores of nodes in internal layers ~\cite{10.1371/journal.pone.0130140}. 

\hspace{3mm} Class Activation Mapping (CAM) highlights class-specific discriminative regions using weighted linear sum of the activation maps \cite{zhou2016learning}. To obtain class activation maps, predicted class scores are mapped with the activation maps of the last convolutional layers, and then all values are summed. When the last activation maps are usually smaller than input images, class activation maps are upsampled to the size of the input images. CAM is applied to time series classification previously \cite{wang2017time}. 

\hspace{3mm} Network Dissection is a method of quantifying the interpretability of individual units in a deep CNN \cite{bau2017network}. This method evaluates the degree of matching of individual units and corresponding concepts from segmentation labels. At this time, the match between a unit and a concept is quantified through the Intersection over Union score (IoU). Therefore, this approach is not applicable to datasets without segmentation labels, such as most time series data.

\begin{table*}[h]
\caption{Comparison with related works and our framework.ND* stands for Network Dissection.}
\label{Comparison with related works}
\vskip -0.1in
\begin{center}
\begin{small}
\begin{sc}
\begin{tabular}[width=.8\linewidth]{lccccc}
\toprule
 & LRP & CAM & ND & CPHAP\\
\midrule
Channel-specific analysis & X & X & O & O\\ 
\midrule
Require Manuel Labels & X & X & O & X\\ 
\midrule
Description method & point by point & point by point & sub-input & sub-input\\ 
\bottomrule
\end{tabular}
\end{sc}
\end{small}
\end{center}
\vskip -0.1in
\end{table*}

\section{Clustered Pattern of Highly Activated Period (CPHAP)}
\begin{figure}[h]
    \includegraphics[width=\linewidth]{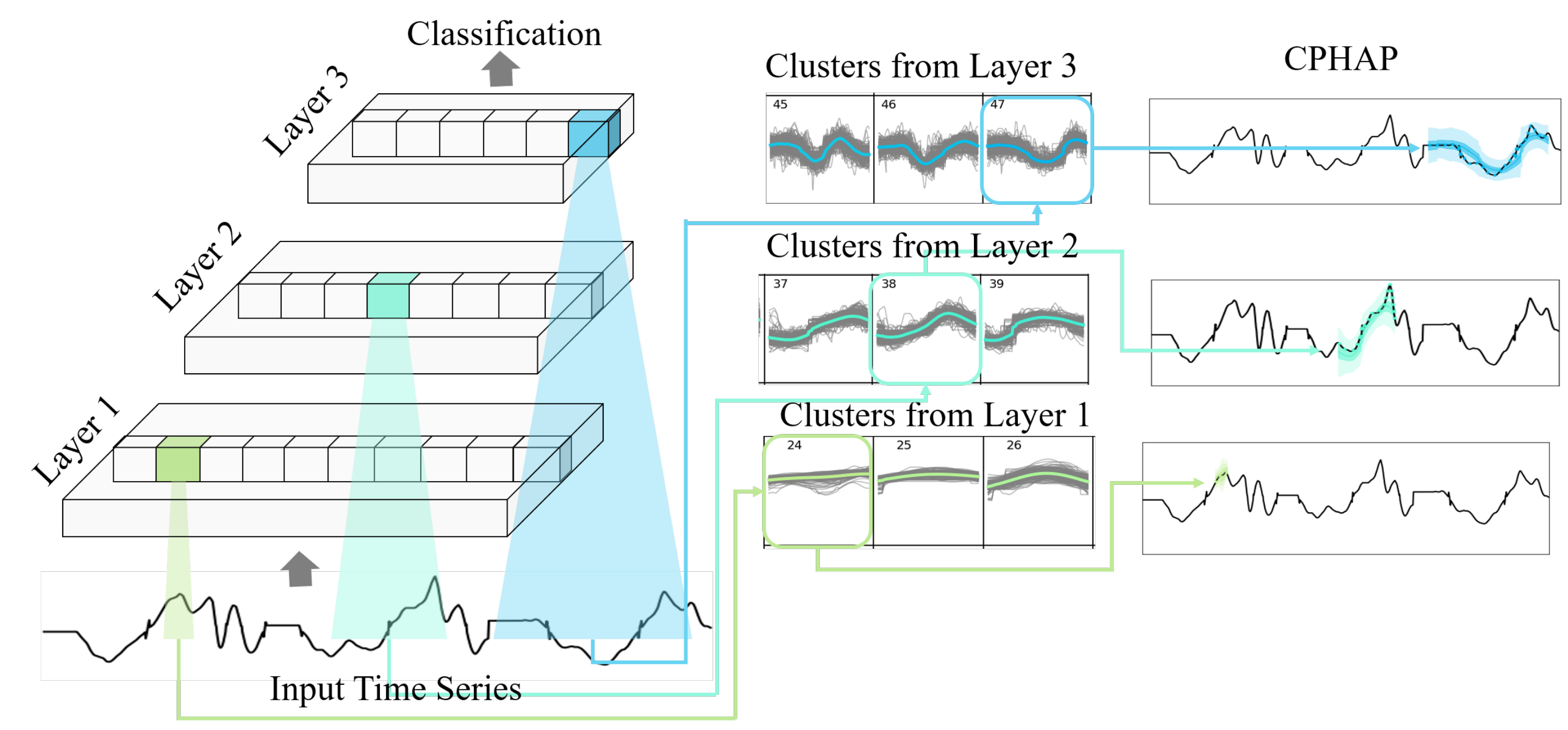}
    \caption{The process to Apply CPHAP}
    \label{fig: Model_structure}
    \vspace{-5mm}
\end{figure}

\subsection{Convolutional Neural Networks}

\hspace{3mm} Convolutional Neural Networks (CNNs) ~\cite{lecun1998gradient} can learn spatial or temporal information from data. The decision process of classification models based on CNNs can be divided into two parts: extracting the features of the input considering spatial correlation and classifying the data into the classes according to these features. Filters in CNNs learn to extract appropriate features from input data during the training procedure. In the inference process, channels from different learned filters reflect the features of the input with activation maps. 

\hspace{3mm} This approach can also be utilized to train models for time series data. Because data points in each subset are highly correlated, learning time series data requires the extraction of local information from each subset of data. Neural Networks which share weight parameters over temporal steps including Recurrent Neural Networks (RNNs), Long-Short Term Memory (LSTM) \cite{greff2016lstm} and Gated Recurrent Unit (GRU) \cite{chung2014empirical} hav been successfully used in natural language processing where data of interest has discrete values. However, CNN-based methods, such as ST-GCN \cite{yan2018spatial} and WaveNet \cite{oord2016wavenet}, have demonstrated outstanding performance on classification and regression tasks of time series data with continuous values. Furthermore, analyzing the role of hidden units in a RNN is difficult and complicated due to RNN's recursive structure. Therefore, we choose CNN-based models to figure out the roles of channels.

\hspace{3mm} In a convolutional layer, a partial region of each channel receives information from only a restricted subset of the output of previous layers. This input subset is called a receptive field. The deeper the layer has, the longer the receptive field in the input. In the next step, we choose critical receptive fields among the entire set of candidates; these receptive fields represent what the channels in CNNs capture from the input.

\subsection{Extracting important input sub-sequences}

\hspace{3mm} We think that analyzing highly activated parts of activation maps play an important role to understand temporal patterns in neural networks. So, our algorithm interprets the decisions of neural networks by extracting highly activated nodes in a channel\footnote{Network Dissection and CAM call this channel as "unit". In this paper, we need to distinguish between a channel and the basic elements in a channel, so we call an activation map as a "channel" and an element of channel as a "node".} and visualizing sub-sequences of the input data that contribute to the highly activated nodes.

\textbf{Highly Activated Period (HAP)}: {\it Highly Activated} means that certain nodes in a channel have bigger values than the channel's threshold. We calculate a threshold $T_{j,k}$ satisfying $P(a_{j,k} > T_{j,k}) = 0.05$ to select the highly activated nodes for each channel $k$ at layer $j$, where $T_{j,k}$ is a threshold of channel $k$ at hidden layer $j$, and $a_{j,k}$ is the distribution of the $k$th channel activations at hidden layer $j$. 

Then, we define Highly Activated Nodes (HAN) as a set of nodes of channel $k$:
\begin{equation}
\label{equ:2}
\textrm{\textbf{HAN}}_{j,k} = \{ i \in [1, N_{j,k}]\;|\; A_{j,k}[i] > T_{j,k} \}
\end{equation} 
where $i$ is a node in channel $k$, $N_{j,k}$ is the number of nodes in channel $k$ at hidden layer $j$ and $A_{j,k}[i]$ is the activation value of a node $i$. The union of the input receptive fields of nodes in HAN is \textbf{Highly Activated Period (HAP)}:
\begin{equation}
\label{equ:3}
\textrm{\textbf{HAP}}_{j,k} = \{ \textrm{Temporal indices in input receptive field}(i) \;|\; i\in \textrm{\textbf{HAN}}_{j,k}\} 
\end{equation}

These periods are the important sub-sequences that we are looking for. 

\begin{algorithm}[h!]
    \caption{Extract important input sub-sequences}
    \label{alg:pattern}
\begin{algorithmic}
    \STATE {\bfseries Input:} data $\mathbf{X}$, Trained CNN model
    
    \FOR {$j$}
        \FOR {$k$}
            \STATE Compute $a_{j,k}$ = the distribution of the $k$th channel activations at hidden layer $j$ in the CNN model for data $\mathbf{X}$
            \STATE Find a threshold $T_{j,k}$ satisfying $P(a_{j,k} > T_{j,k}) = 0.05$
        \ENDFOR
    \ENDFOR
 
    \STATE Define $\mathrm{HAP\_list}_{j}$ = [ ] for all $j$
    \FOR {$x\in \mathbf{X}$}
        \FOR {$j$}
            \FOR {$k$}
                \STATE $A_{j,k}$ = the activation values of the $k$th channel at hidden layer $j$ in the CNN model  
                \STATE $\textrm{\textbf{HAN}}_{j,k}(x) = \{ i \in [1, N_{j,k}]\;|\; A_{j,k}[i] > T_{j,k} \}$ where $i$ is a node in channel $k$
                \STATE $\textrm{\textbf{HAP}}_{j,k}(x) = \{ \textrm{Temporal indices in input receptive field}(i) \;|\; i\in \textrm{\textbf{HAN}}_{j,k}(x)\}$
                \STATE $\mathrm{HAP\_list}_{j}$.append($\textrm{\textbf{HAP}}_{j,k}(x)$)
            \ENDFOR
        \ENDFOR
    \ENDFOR
    \STATE Train SOM clustering with HAP\_list given the number of clusters
    \STATE Choose data $x$, layer $j$, channel $k$
    \STATE Compute $p = \textrm{\textbf{HAP}}_{j,k}(x)$
    \STATE Compute cluster $c_{p}$ = SOM($p$)
    \STATE $\textbf{\textrm{CPHAP}}_{p}$ = mean(\{$p'\in\mathrm{HAP\_list}_{j}\;|\;$SOM($p'$) = $c_{p}$\})
    \STATE $\textbf{\textrm{UNCERTAINTY}}_{p}$ = variance(\{$p'\in\mathrm{HAP\_list}_{j}\;|\;$SOM($p'$) = $c_{p}$\})
\end{algorithmic}
\end{algorithm}

\subsection{Patternizing representative sub-sequences by clustering }

\hspace{3mm} So far, we have found important sub-sequences to figure out what the individual channels in convolutional neural networks are looking at in the time series data. Now, we characterize these sub-sequences and assign the general shapes to them. In this paper, we use Self Organizing Map (SOM) to characterize temporal patterns in a HAP \footnote{We experimentally try various clustering methods, including comparing several clustering methods, including K-means clustering, Gaussian Mixture Models (GMMs), K-shape clustering \cite{kshape} and Self Organizing Map (SOM). The outputs of each method are provided in the appendix.}.

\hspace{3mm} SOM is an unsupervised learning method by mapping high-dimensional data to a low-dimensional map. The weight vectors in the map are initialized to random values in the high-dimensional input space. During the training procedure, weight vectors move toward the input data with keeping the topology of the map space. The trained weight vectors works as cluster groups. We use a $8\times8$ map, so a total of 64 groups show various patterns and each group has low variance among elements in the same group. Furthermore, SOM represents the relationship among cluster groups because near cluster groups show similar patterns on the map. It is shown in Figure 2. After clustering, we compute an average of each cluster group to assign a pattern.

 \textbf{CPHAP}: A Clustered Pattern of Highly Activated Period is an average over time axis of a cluster group $C$ given \textrm{HAP\_list} which means a list of temporal sequences that activates nodes. 
\begin{equation}
\label{equ:4} 
    \textrm{\textbf{CPHAP}} = mean(\{p\in \textrm{HAP\_list}\;|\;SOM(p) = C\})
\end{equation}
Note that HAP\_list should be defined by layer, since channels in different layer have different lengths of input receptive field. 

\hspace{3mm} Our framework also illustrates the uncertainty in CPHAP. The uncertainty in CPHAP is the variance for the posterior probability of the corresponding the input sub-sequences. It indicates how much certainty the input sequences belongs to a cluster. Input sub-sequences, that appear frequently in a set of the receptive field of the spatial region of the highly activated unit and have distinctive shapes, belongs to an less uncertainty of cluster. In contrast, sub-sequences, nor have  characteristic shape neither appear on the main input, belongs into the rest cluster which have very large variance clusters. The entire process is explained in Algorithm 1.  

\begin{figure}[h]
    \includegraphics[width=\linewidth]{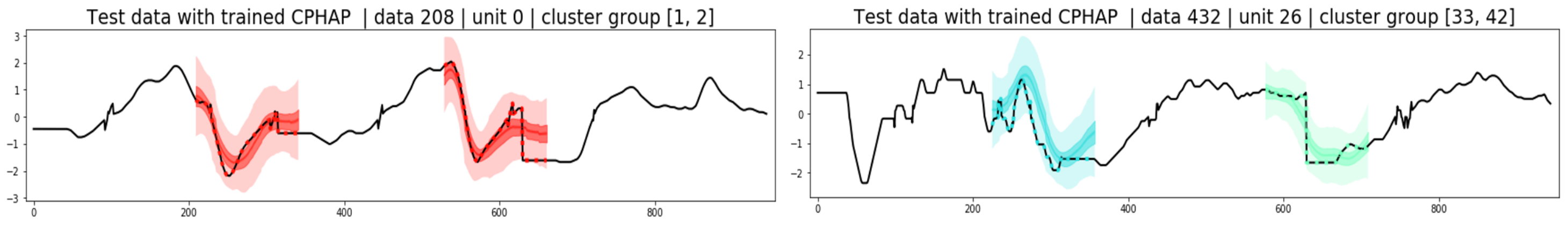}
    \caption{CPHAPs are also well matched to new test data. These sequences are from the test dataset which did not be used for getting CPHAP, while CPHAPs are from the training dataset.}
    \label{fig: test_horizon}
  \vspace{-5mm}
\end{figure}

\section{Experimental Results}
\textbf {Dataset}
We use three open time series dataset for experiments; \textit{UWaveGestureLibraryAll}~\cite{LIU2009657} is a set of eight simple gestures generated from accelerometers, \textit{Smartphone Dataset for Human Activity Recognition (HAR)}~\cite{Dua:2019} is a smartphone sensor dataset recording human perform eight different activities. 

\textbf {Model}  
We use a temporal CNNs which is composed of three convolution layers followed by pooling layers, and one fully connected layer. ReLU function is used as the activation function in every hidden layer. Batch size and epoch are 64 and 500 respectively. In the prediction phase, we choice the model with the lowest validation loss. Even we mainly use this setting for our experiments, we also apply to ResNet as shown in Figure 4 \cite{wang2017time}.

\begin{figure}[h]
  \includegraphics[width=\textwidth]{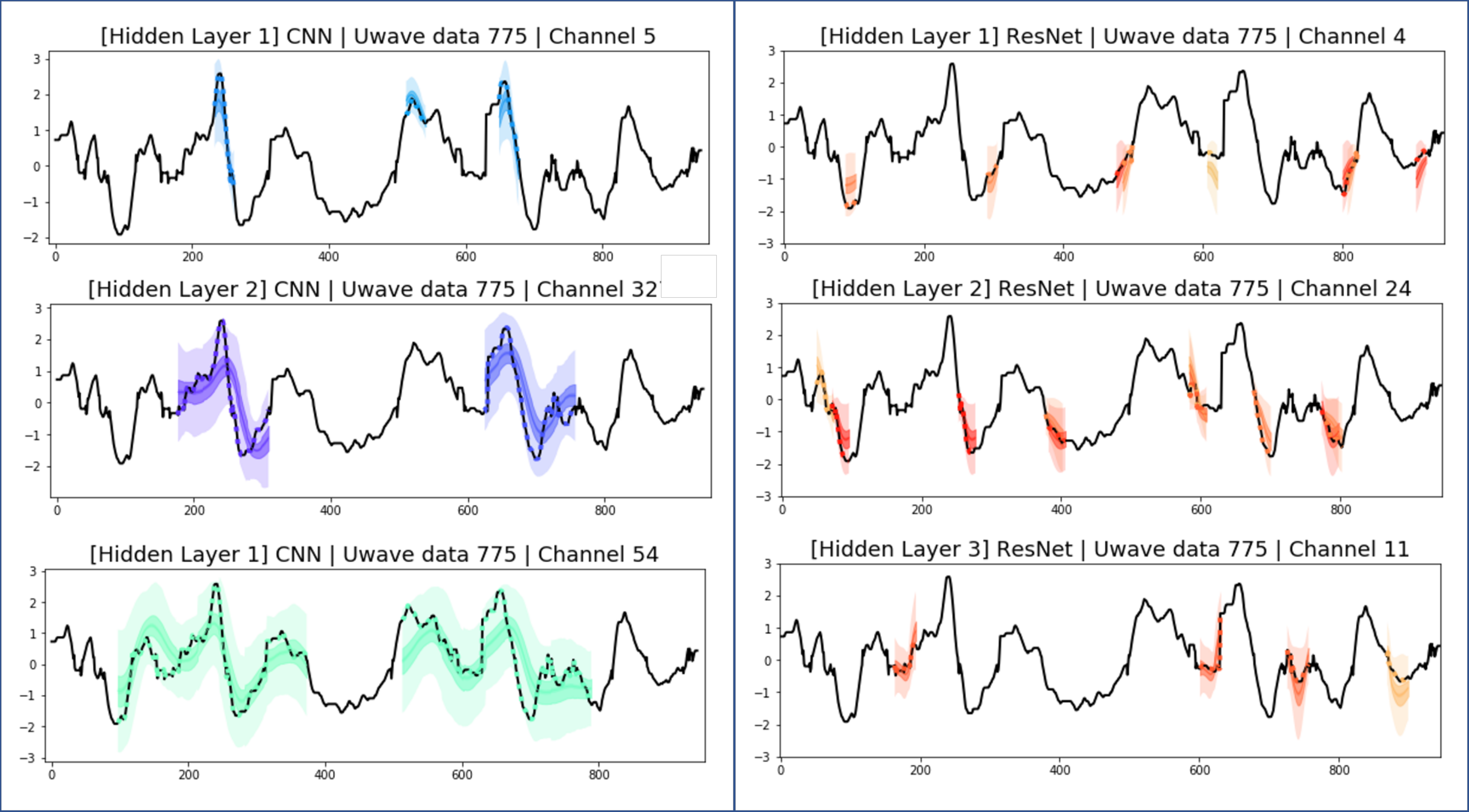} 
  \caption{ For ResNet, each hidden layer has a different length of CPHAP; 15, 25, 31, 45, 55, 61, 75, 85 and 91. The pattern lengths are depending on model structures like kernel size, pooling size, or stride size, etc. }
  \label{fig:resnet}
\end{figure}


\subsection{Identifying role of the channel in CPHAP}

\begin{figure}[h]
    \includegraphics[width=\linewidth]{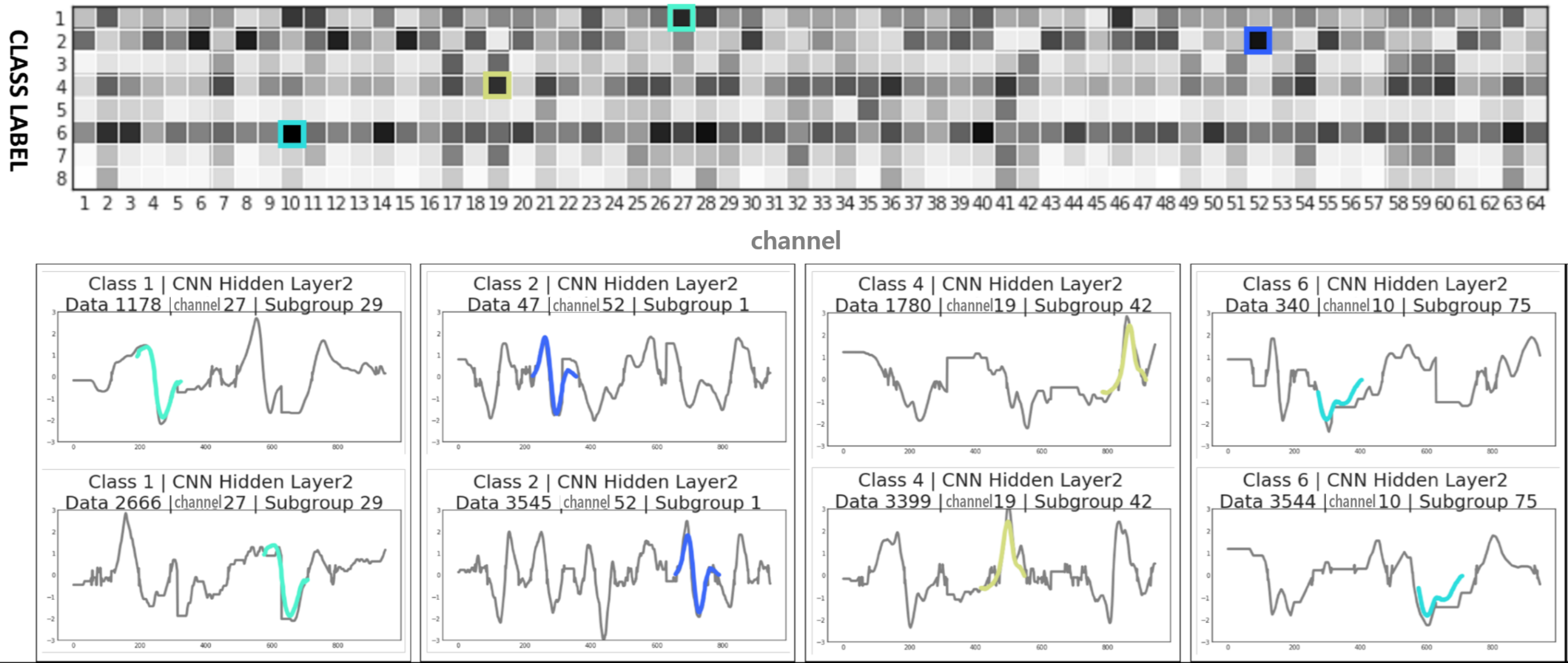}
    \caption{(Top) Highly related class-unit pairs are (Class1-Channel27) , (Class2-Channel52), (Class4-Channel19) and (Class6-Channel10) in EEG Dataset. (Bottom) Visualization of the 4 most distinguishable Class-Channel relationships for Uwave with the color of the pattern that is mostly activated the corresponding unit.}
    \label{fig:UWAVE_class_pattern}
\end{figure}

\hspace{3mm} To analyze the relationships between channel and class labels, we perform the following steps.
\begin{enumerate}
\item $freq_{U,P}$  : Count patterns $p$ contributing to activating channels $u$ for every $u$ and $p$ pair.
\item $freq_{P,C}$  : Count patterns $p$ belonging in the class $c$ for every $p$ and $c$ pair.
\item $freq_{U,C}$  : Multiply $freq_{U,P}$ and $freq_{P,C}$.
\end{enumerate}
 The value of  $freq_{U,C}\left[ u,c \right]$ indicates how much unit $u$ pays attention to frequent pattern in class $c$. Figure ~\ref{fig:UWAVE_class_pattern} is the frequency matrix $freq_{u,c}$ from the second layer of CNN with Uwave dataset and it visualizes the 4 most distinguishable Class-Channels relationships with the color of the patterns that is mostly activated the corresponding channel. The first columns show the relationships with Class 1 and Channel 27. Channel 27 is mainly activated by a sub-sequence similar to an inverted concave shape, frequently seen in class 1. In the second column, channel 52 is activated by the concave and convex patterns, so we can infer that this activation of Channel 52 plays a major role to classify whether this input belongs to class 2. In the third column, the role of channel 19, which primarily looks like a convex shape, stands out when deciding class 4. The fourth column shows channel 10 is activated by an alphabet 'W' shape to predict class 6.  The outputs of other dataset are in appendix.

\subsection{ Comparisons with other interpretable methods}

\hspace{3mm} We try to compare our framework with LRP, since LRP does not provide channel-specific analysis, we apply LRP to highly activated nodes of a specific channel which we call "Channel-LRP". Figure 6 depicts how each method interpret and visualize the role of the channel in neural networks. Colorized points in Channel-LRP show importance of each point from the input with color variation. However, the results of Channel-LRP are not fully reliable for users to accept the focused parts as continuous sub-sequences because they are represented as points apart. On the other hand, CPHAP provides clear explanation for the role of the channel and visualizes which pattern each sub-sequence is perceived as in neural networks. Thus, users can recognize the visualized results as semi-global shape intuitively. Furthermore, our framework shows the uncertainty for each pattern, which implies the potential range of inputs activating corresponding channel. Note that each time point in the pattern has different degree of uncertainty.

\begin{figure}[h]
  \includegraphics[width=\textwidth]{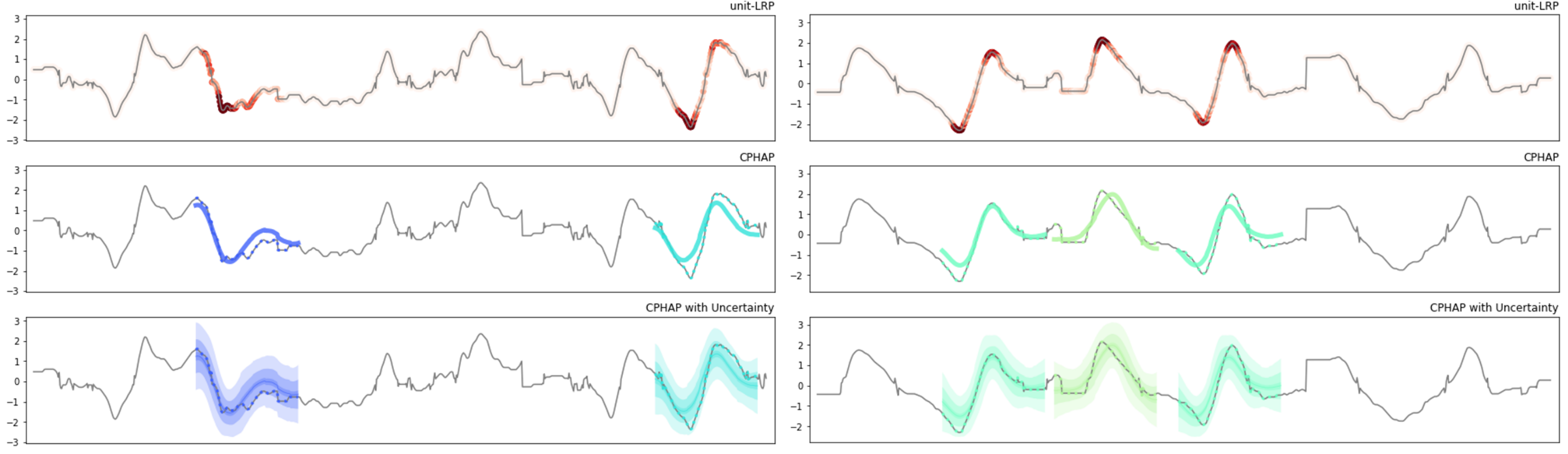}
  \caption{Comparison with other Interpretable methods; (a) channel-LRP, (D) CPHAP, (E) CPHAP with uncertainty  }
  \label{fig:teaser}
\end{figure}

\subsection{Perturbation analysis}

\begin{figure}[h]
    \includegraphics[width=\linewidth]{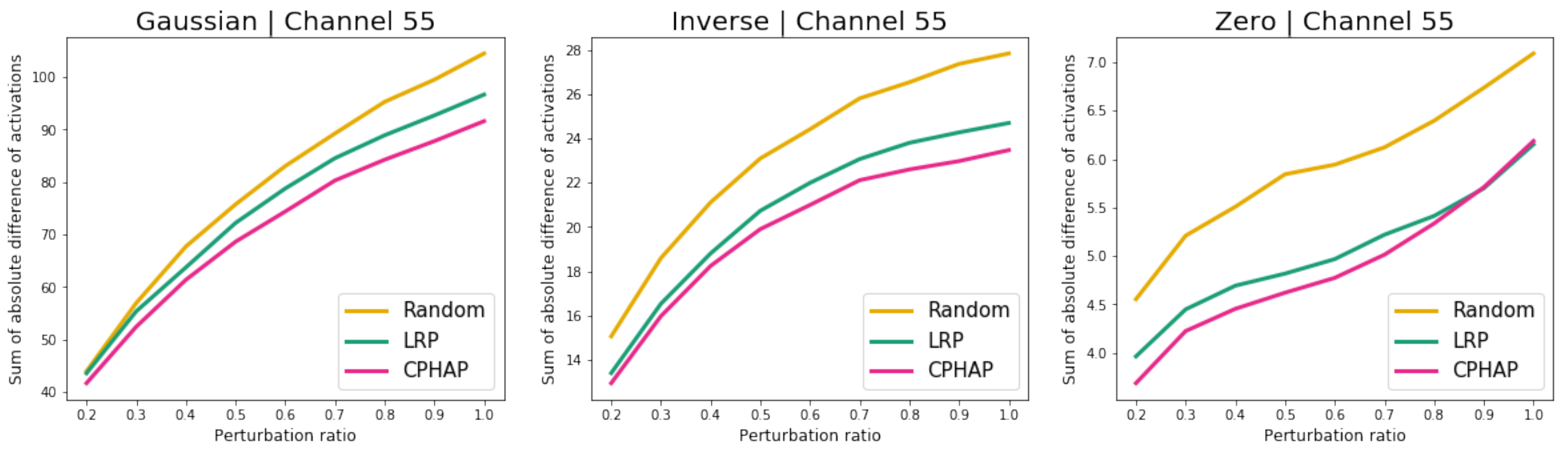}
    \caption{This graph illustrates the results of perturbations on the input while preserving the regions selected by each method. We apply three kinds of perturbations: Gaussian perturbation, Inverse perturbation and zero perturbation. The x-axis means the ratio of perturbed regions except preserved regions. The y-axis means the sum of changes of activations in the channel 55.}
    \label{fig:turn_off}
\end{figure}

\begin{table*}[ht!]

\begin{tabularx}{\linewidth}{l|XXXXX}
\toprule
Gaussian     & 0.2              & 0.4              & 0.6              & 0.8              & 1.0                \\
\midrule
vs Random & 10.06$\pm$7.08\% & 11.02$\pm$7.21\% & 11.4$\pm$7.15\%  & 11.68$\pm$7.43\% & 12.4$\pm$7.38\%  \\
vs LRP    & 5.72$\pm$3.96\%  & 4.97$\pm$2.73\%  & 4.92$\pm$2.04\%  & 4.83$\pm$1.88\%  & 5.11$\pm$1.91\%  \\
\bottomrule
\end{tabularx}
\vskip 0.05in

\begin{tabularx}{\linewidth}{l|XXXXX}
\toprule
Inverse     & 0.2              & 0.4              & 0.6              & 0.8              & 1.0                \\
\midrule
vs Random & 17.28$\pm$6.89\% & 11.24$\pm$9.37\% & 10.44$\pm$10.91\% & 12.31$\pm$10.42\% & 15.03$\pm$9.1\% \\
vs LRP & 4.93$\pm$3.72\% & 0.41$\pm$3.73\% & 0.26$\pm$3.61\% & 1.28$\pm$3.44\% & 2.45$\pm$2.86\% \\
\bottomrule
\end{tabularx}
\vskip 0.05in

\begin{tabularx}{\linewidth}{l|XXXXX}
\toprule
Zero     & 0.2              & 0.4              & 0.6              & 0.8              & 1.0                \\
\midrule
vs Random & 20.27$\pm$6.08\% & 19.48$\pm$6.41\% & 19.29$\pm$6.99\% & 19.96$\pm$7.02\% & 21.2$\pm$6.92\% \\
vs LRP & 5.85$\pm$3.5\% & 2.09$\pm$4.13\% & 0.33$\pm$4.23\% & -0.38$\pm$3.5\% & -0.41$\pm$3.23\% \\
\bottomrule
\end{tabularx}
\caption{The Robustness of internal activations when applying perturbations while preserving sub-sequences selected by CPHAP. The values in the table mean how less activations in channels change with CPHAP in perturbation compared to other methods (mean$\pm$ std). The first row shows the type of the perturbation and the perturbation rates. In most cases, CPHAP has better or similar performance compared to LRP.}
\end{table*}

\hspace{3mm}  Our framework has successfully identified important sub-sequences in which each channel focuses from time series input data. Now we evaluate the effects of these sub-sequences on the corresponding channel quantitatively. We assume that the activation values in the channel are robust to perturbations in the input with preserving the detected important sub-sequences.

\hspace{3mm}  First, we randomly choose regions to apply perturbations while preserving specific regions selected by each method. The methods include CPHAP, LRP and Random. We observed changes in activation values of given channels with different perturbation rates on the input. In the case of LRP, we calculate relevance scores and time points which have high relevance scores are selected to be preserved. The number of selected time points is equal to the case of CPHAP. No points are preserved when we use Random method.

\hspace{3mm}  After randomly choosing points from the input data except preserved period, we apply perturbations and observe changes in activations of the channel. We apply three kinds of perturbations: Gaussian, Inverse, Zero. Gaussian perturbation replaces the points with randomly sampled values from $Normal(0,3)$. Inverse perturbation reverses the sign of points and Zero perturbation assigns zero values to the points. These noisy inputs are forward-propagated to the model in order to compute activation values in the given channel. For each channel, we calculate the absolute values of activation difference after perturbations. Figure 7 shows the result in the channel 55 of layer 2 on Uwave dataset.

\hspace{3mm} Table 2 shows how the results from CPHAP are less vulnerable to perturbations than the other methods on Uwave dataset. Our performance is always better than Random and LRP when using Gaussian perturbations. Though activation difference from CPHAP is slightly worse than LRP when using Zero perturbation with ratio of 0.8 and 1.0, the degree of differences are very close to zero and CPHAP surpasses LRP method in most of the other cases. It implies that CPHAP can select critical regions from the input better than the other methods and those selected regions represent the role of the channel in CNN models.

\section{Conclusion} 
\hspace{3mm} We propose a new method to visualize the representative temporal patterns in neural networks by breaking into the activation of channels and analyzing their receptive fields in the input space. Our framework uses highly activated temporal regions and applies clustering method to patternize these regions in order to obtain the general shapes without human-segmented information. Also, we calculate the uncertainty of each cluster, which enables users to gain insight into the type of the new input. Our work breaks down interpretation of deep neural networks for the internal activation into channel-by-channel roles for easy visualization, and help users to interpret the temporal neural network intuitively. In addition, our perturbation analysis shows that temporal regions are useful to analyze contribution to node activations than points selected by relevance scores in time series data.



\begin{ack}
 This work was supported by IITP grant funded by the Korea government(MSIT) (No.2017-0-01779, XAI) 
\end{ack}

\bibliographystyle{acm}
\bibliography{main}

\newpage
\section{Broader Impact}
 \hspace{3mm} Our framework provides the interpretation of deep neural networks for the internal activation into channel-by-channel roles with easy visualization, and helps users to understand the decision-making process of temporal neural network intuitively. Our framework automatically extracts the representative temporal patterns from time series data without human-annotated segmentation information, which is difficult to obtain from actual time series data. Thus, CPHAP could be applied as the source technology to various tasks that require temporal segmented information and transparency of AI systems. Since the amount of time series data would increase significantly in future industries, our work would inspire people to understand AI systems and apply useful AI technologies to real world problems.

\section{Experimental Settings}
\begin{table*}[h]
\caption{Datasets Information}
\label{Datasets Information}
\vskip 0.05in
\begin{center}
\begin{small}
\begin{sc}
\begin{tabular}{cccccc}
\toprule
      & Time Length & Input Sensor & Class & Training Data & Test Data  \\
\midrule
UWave & 945         & 1            & 9      &3582       &896  \\
HAR   & 128         & 9            & 6      &7352      &2947 \\
\bottomrule
\end{tabular}
\end{sc}
\end{small}
\end{center}
\vskip -0.1in
\end{table*}

\begin{table*}[h]
\caption{ CNN Structures used in our framework }
\label{CNNStructures}
\vskip 0.05in
\begin{center}
\begin{small}
\begin{sc}
\begin{tabular}[width=\textwidth]{lcccccccccc}
\toprule
\shortstack{dataset}  &\shortstack{Test\\ Accuracy} &
\shortstack{conv1\\filter} &
\shortstack{pool1\\size} &  \shortstack{conv2\\filter} & \shortstack{pool2\\size} & \shortstack{conv3\\filter} & \shortstack{pool3\\size} \\
\midrule
UWave  & 98.2\% &15 & 8 & 11 & 8 & 7 & 4 \\ 
HAR & 79.0\% &7 & 4 & 5 & 4& 3 & 2 \\ 
\bottomrule
\end{tabular}
\end{sc}
\end{small}
\end{center}
\vskip -0.1in
\end{table*}

\begin{table*}[h]
\caption{ ResNet Structures used in our framework}
\label{ResNet Structures}
\vskip 0.05in
\begin{center}
\begin{small}
\begin{sc}
\begin{tabularx}{\linewidth}{llXXXXXXXXX}
\toprule
\shortstack{ }&  \shortstack{ test\\ accracy }  &\shortstack{ conv1\\filter}  &\shortstack{ conv2\\filter }  &\shortstack{ conv3\\filter }  &\shortstack{ conv4\\filter }  &\shortstack{ conv5\\filter }  &\shortstack{ conv6\\filter }  &\shortstack{ conv7\\filter }  &\shortstack{ conv8\\filter }  &\shortstack{ conv9\\filter }\\
\midrule
UWave & 89.3\%        & 15           & 11           & 7            & 15           & 11           & 7            & 15           & 11           & 7            \\
HAR   & 94.9\%        & 8            & 5            & 3            & 8            & 5            & 3            & 8            & 5            & 3           \\
\bottomrule
\end{tabularx}
\end{sc}
\end{small}
\end{center}
\end{table*}

 \vspace{3mm} 

\section{Visualization Methods }
\subsection{ Visualization of CPHAP }
 \hspace{3mm} In this paper, we visualize CPHAPs with colored lines; A black line is an input sequence, a colored dot line is an input receptive field, a colored line is CPHAP and a colored shadow means the uncertainty of the pattern. 
\begin{figure}[h!]
\centering
    \includegraphics[width=\linewidth]{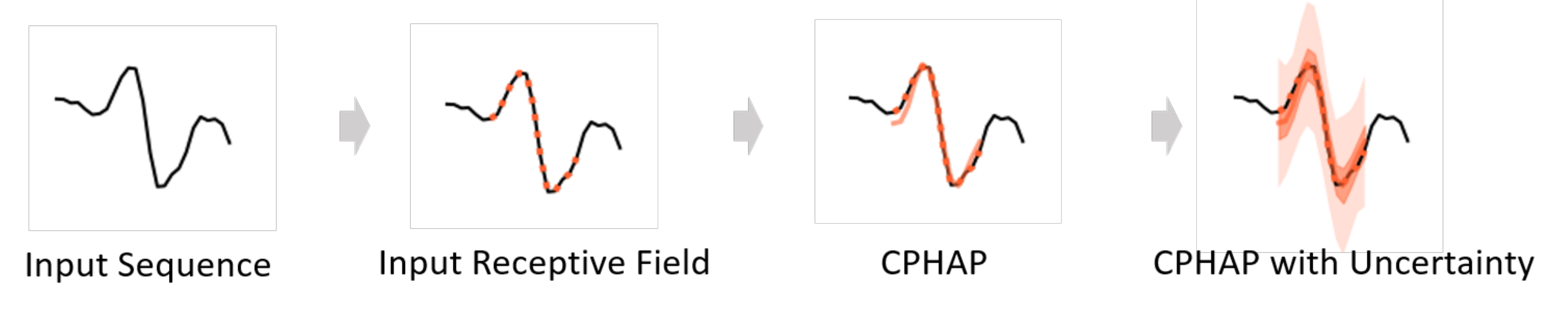}
    \caption{Visualization Process of CPHAP}
    \label{fig: draw_cphap}
\end{figure}

 \hspace{3mm} Figure 2 (a) shows that how the input receptive fields of the channel activations can have the overlapped areas. Then, some parts of patterns detected in receptive fields can also be overlapped with other patterns. To show each pattern more clearly, we remove some overlapped patterns for CPHAP and illustrate the important patterns intuitively.

\begin{figure}[h]
    \centering
    \begin{subfigure}[b]{0.45\textwidth}
        \centering
        \includegraphics[width=\textwidth]{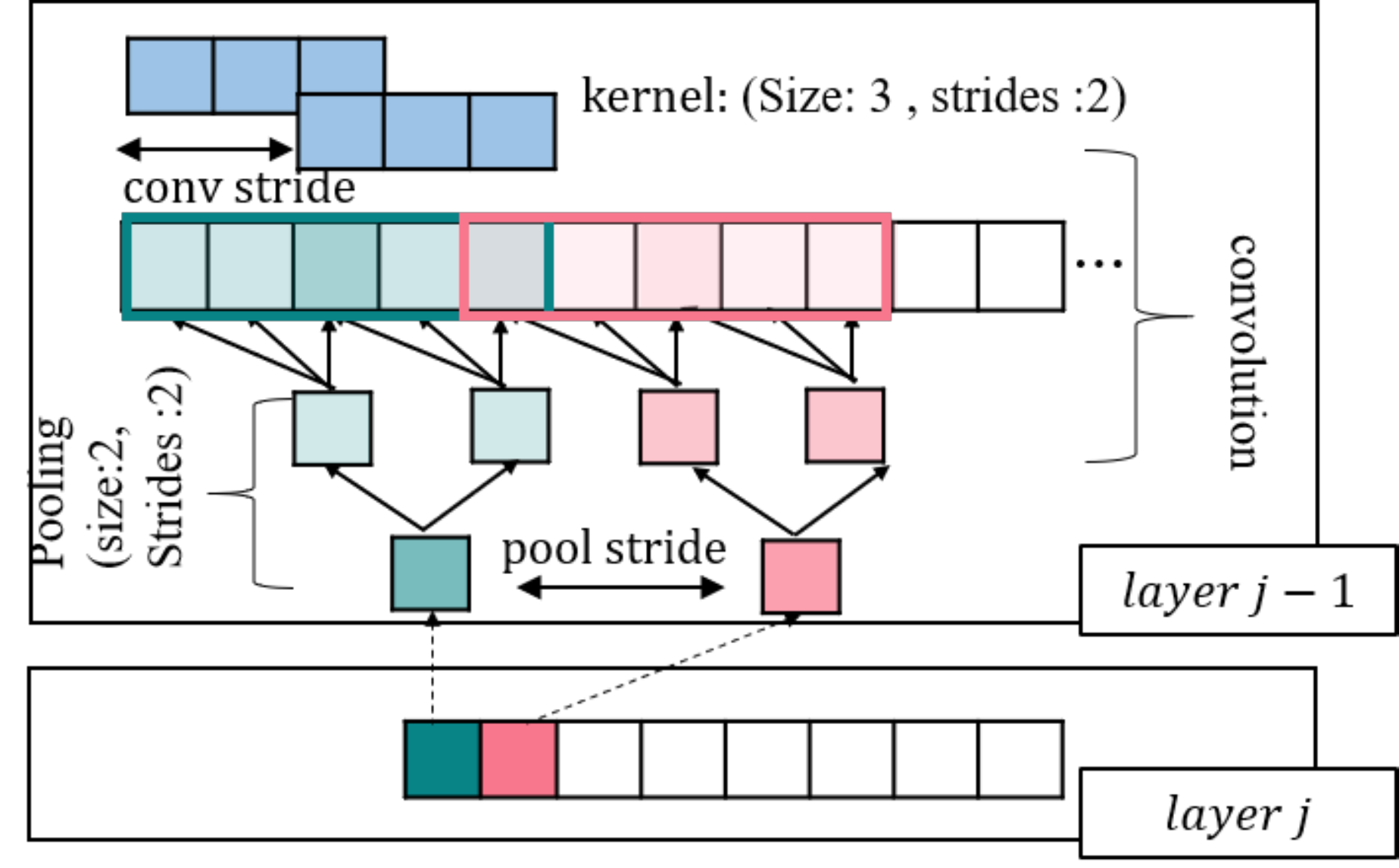}
         \caption{Receptive Field}
    \end{subfigure}
    \hspace{1em}
    \begin{subfigure}[b]{0.45\textwidth}
        \centering
        \includegraphics[width=\textwidth]{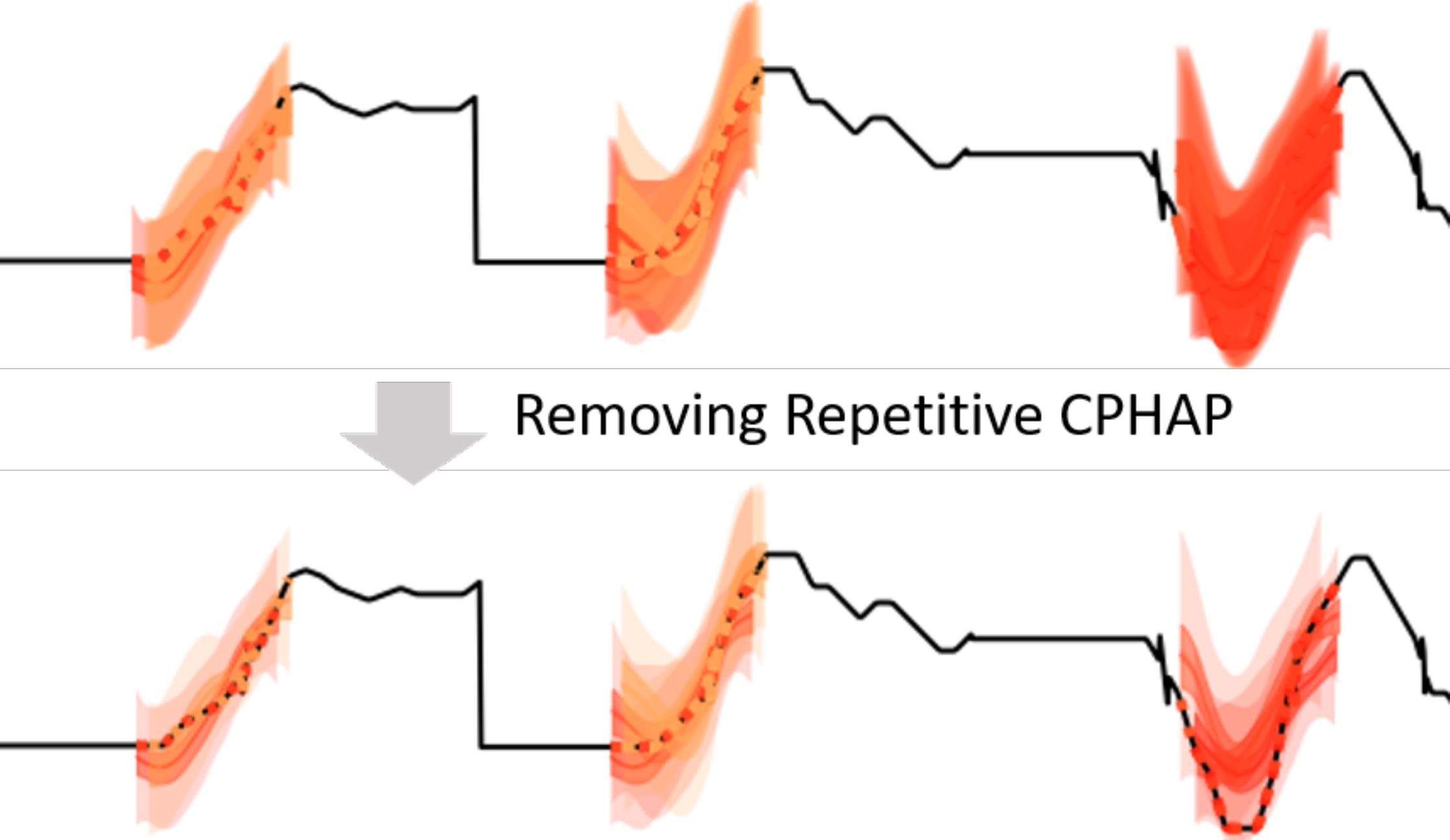}
         \caption{Overlapped CPHAP}
    \end{subfigure}
    \caption{How to Handle Overlapped Patterns}
\end{figure}

\subsection{ Various Methods to Visualize Important Regions }
\hspace{3mm} In Figure 3, we visually compare CPHAP with other methods that interpret neural networks. Since CAM can interpret only the last layer of CNNs and Network Dissection requires pre-defined concepts (e.g. object, texture, color) in order to explain the internal processes of CNNs, it is hard to directly apply these methods to explain internal activations in neural networks for time-series data. However, we can follow these methods in the similar manner. Both CAM and Network Dissection use the activation map to get the important area and upsample it to map toward the original input size. Inspired by this process, we upsample the channel activation first. Then, we highlight the important area with heatmap like CAM or turn off partial regions where the activation value is under the threshold like Network Dissection. The details of Channel-LRP is described in the Section 4.2 of the main paper.

\begin{figure}[h]
    \includegraphics[width=\linewidth]{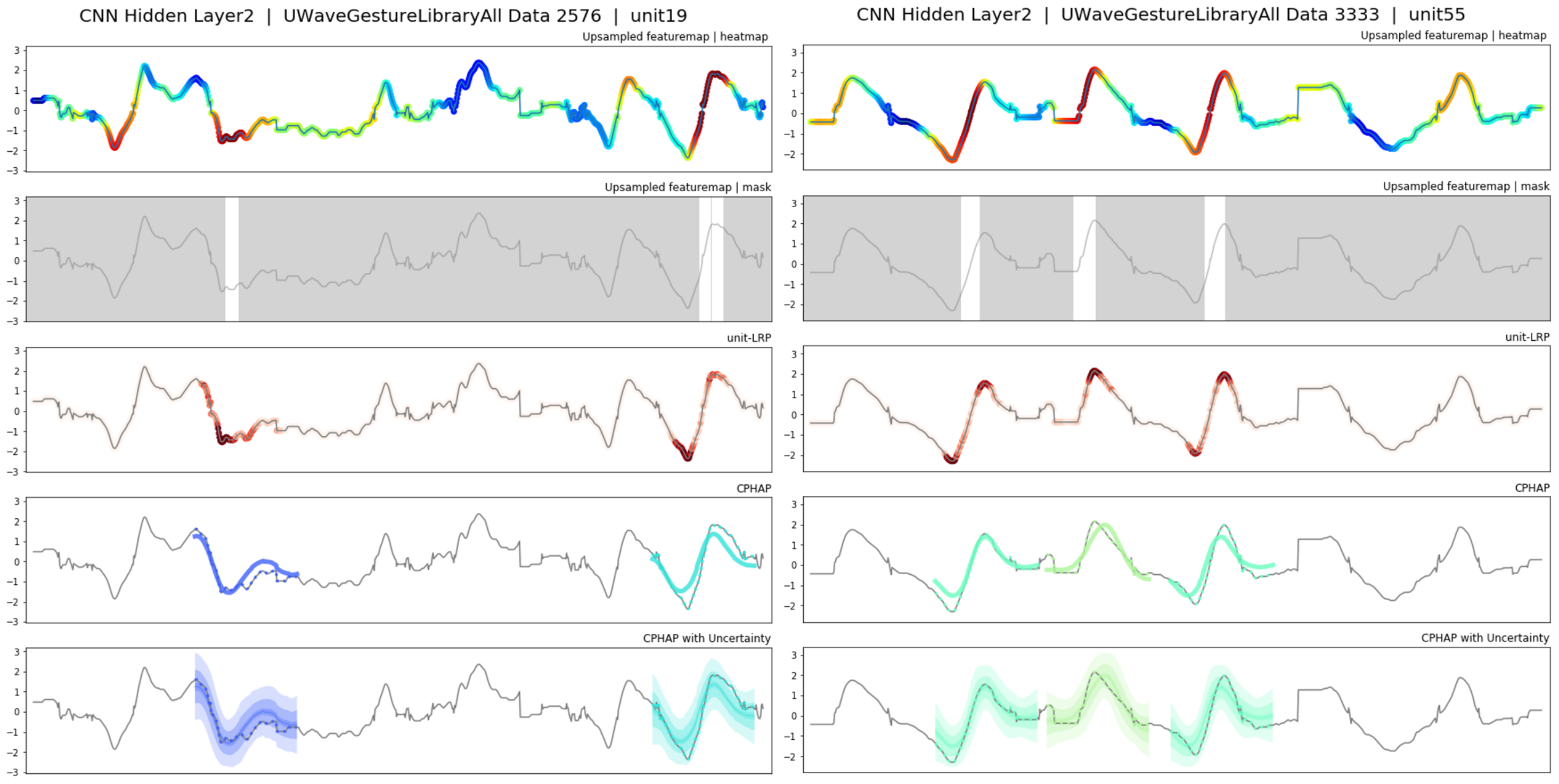}
    \caption{Comparison of Methods that Visualize Important Regions}
    \label{Comparison of Methods}
\end{figure}
\newpage

\section{Additional Experimental Results}
\subsection{ CPHAP Results of ResNet}
 \hspace{3mm} The patterns detected by channels in different layer have different lengths. Patterns from lower layers, such as layer 1, layer 2, and layer 3, reflect local changes like short concave shapes. On the other hands, patterns from higher layers, such as layer 7, layer 8, and layer 9, capture global changes like slow upward trends.
\begin{figure}[h]
    \includegraphics[width=\linewidth]{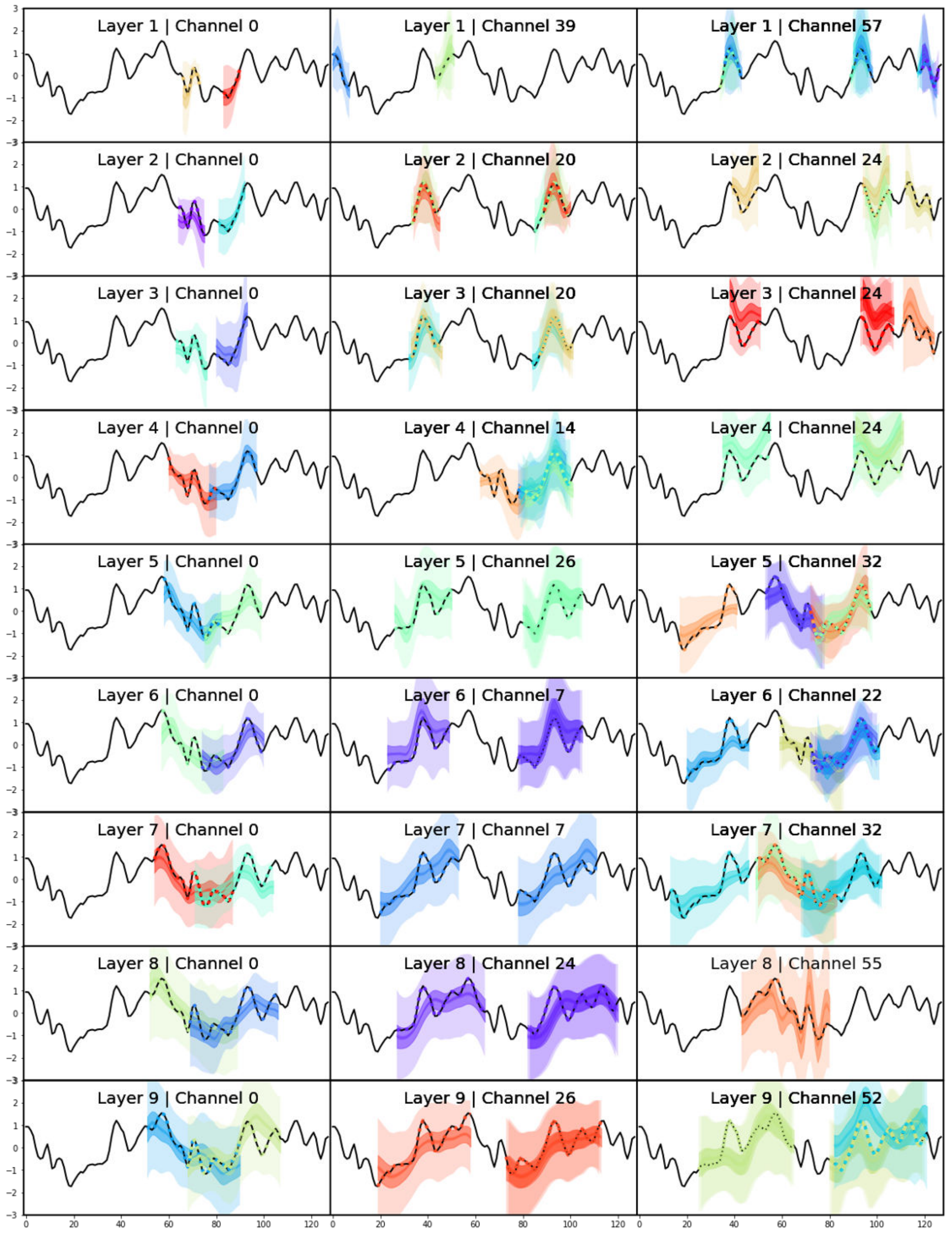}
    \caption{CPHAP Result in ResNet Model, HAR Data 1613, Sensor8}
    \label{fig: resnet_har_1613_1}
\end{figure}
\newpage

 \hspace{3mm} Channel 55 in layer 1 detects sharp decreasing patterns and Channel 51 detects sharp increasing patterns. In layer 4 and layer 5, the channels can capture more complex and longer patterns than lower layers. For example, channel 6 in layer 5 and channel 45 in layer 4 detect a 'W' shape as a pattern. This kind of complex patterns can be smoothed in the higher layer. Actually, this 'W' shape is detected as a 'U' shape by channel 6 in layer 7.

\begin{figure}[h]
    \includegraphics[width=\linewidth]{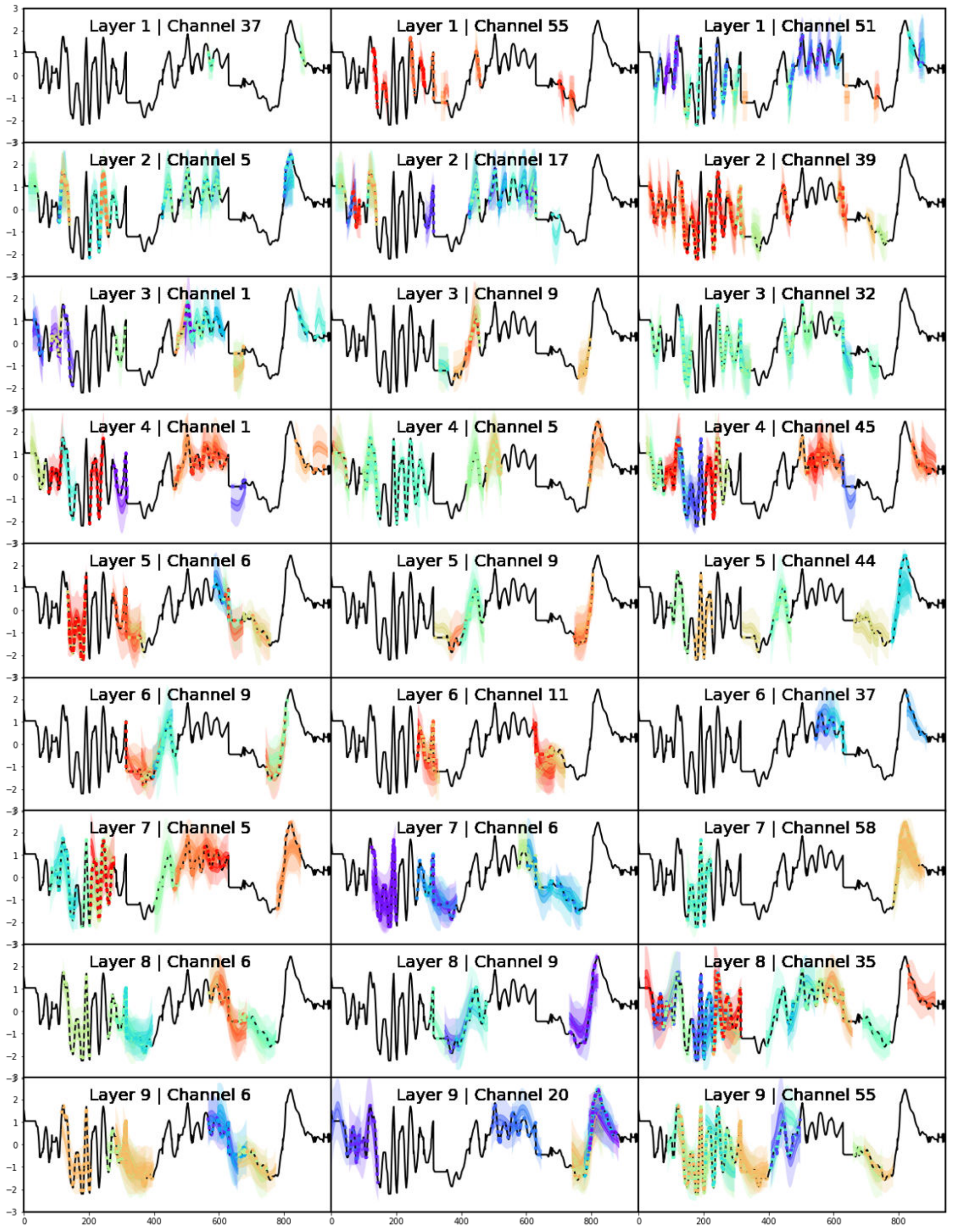}
    \caption{CPHAP Result in ResNet Model, UWave Data 536, Sensor 0}
    \label{fig: resnet_uwave_536_1}
\end{figure}
\newpage

\hspace{3mm} Given the data sample, channels in lower layers tend to focus on rapid changes or inflection points. For this simple sample, even layer 4 and layer 5 do not capture complex patterns. The channels in the higher layers recognize extreme changes in softer patterns.

\begin{figure}[h]
    \includegraphics[width=\linewidth]{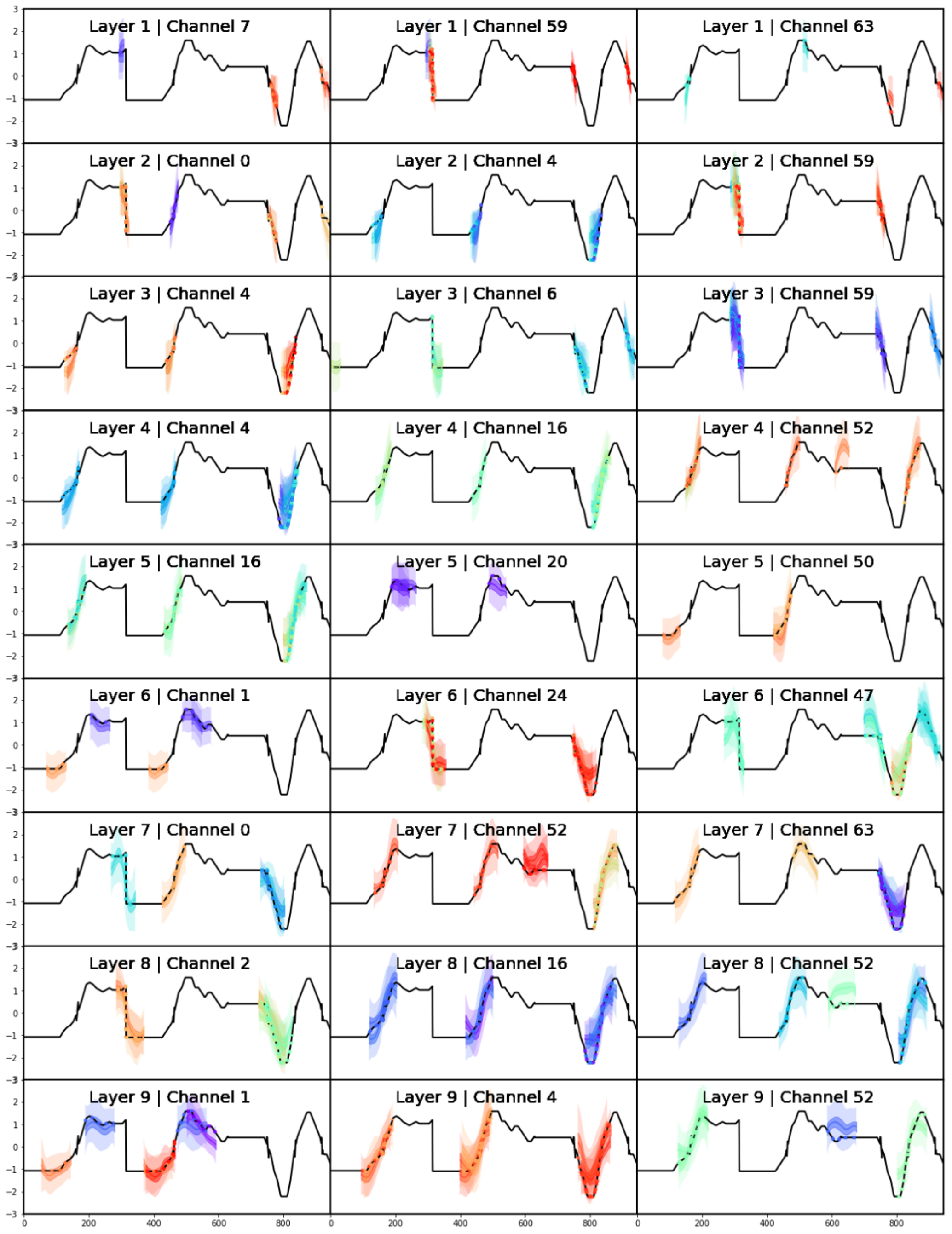}
    \caption{CPHAP Result in ResNet Model, UWave Data 571, Sensor 0}
    \label{fig: resnet_uwave_571_1}
\end{figure}

\clearpage
\subsection{Comparison of Various Filter Sizes}
\hspace{3mm} The longer convolution filters are, the longer patterns are. The longer filter is appropriate for capturing global trends. If a dataset has complex oscillation, it would be better to use shorter filter in order to detect local features.

\begin{table*}[h!]
\caption{ ResNet Structure with Longer Filters}
\label{ResNet Structure with Longer Filters}
\vskip 0.05in
\begin{center}
\begin{small}
\begin{sc}
\begin{tabular}{lccccccccccc}
\toprule
  &\shortstack{ conv1\\filter}  &\shortstack{ conv2\\filter }  &\shortstack{ conv3\\filter }  &\shortstack{ conv4\\filter }  &\shortstack{ conv5\\filter }  &\shortstack{ conv6\\filter }  &\shortstack{ conv7\\filter }  &\shortstack{ conv8\\filter }  &\shortstack{ conv9\\filter }\\
\midrule
filter size  & 15      & 11      & 9       & 15      & 11      & 9       & 15      & 11      & 9       \\
pattern size & 15      & 25      & 33      & 47      & 57      & 65      & 79      & 89      & 97     \\
\bottomrule
\end{tabular}
\end{sc}
\end{small}
\end{center}
\vskip -0.1in
\end{table*}

\begin{figure}[h!]
\centering
    \includegraphics[width=0.85\linewidth]{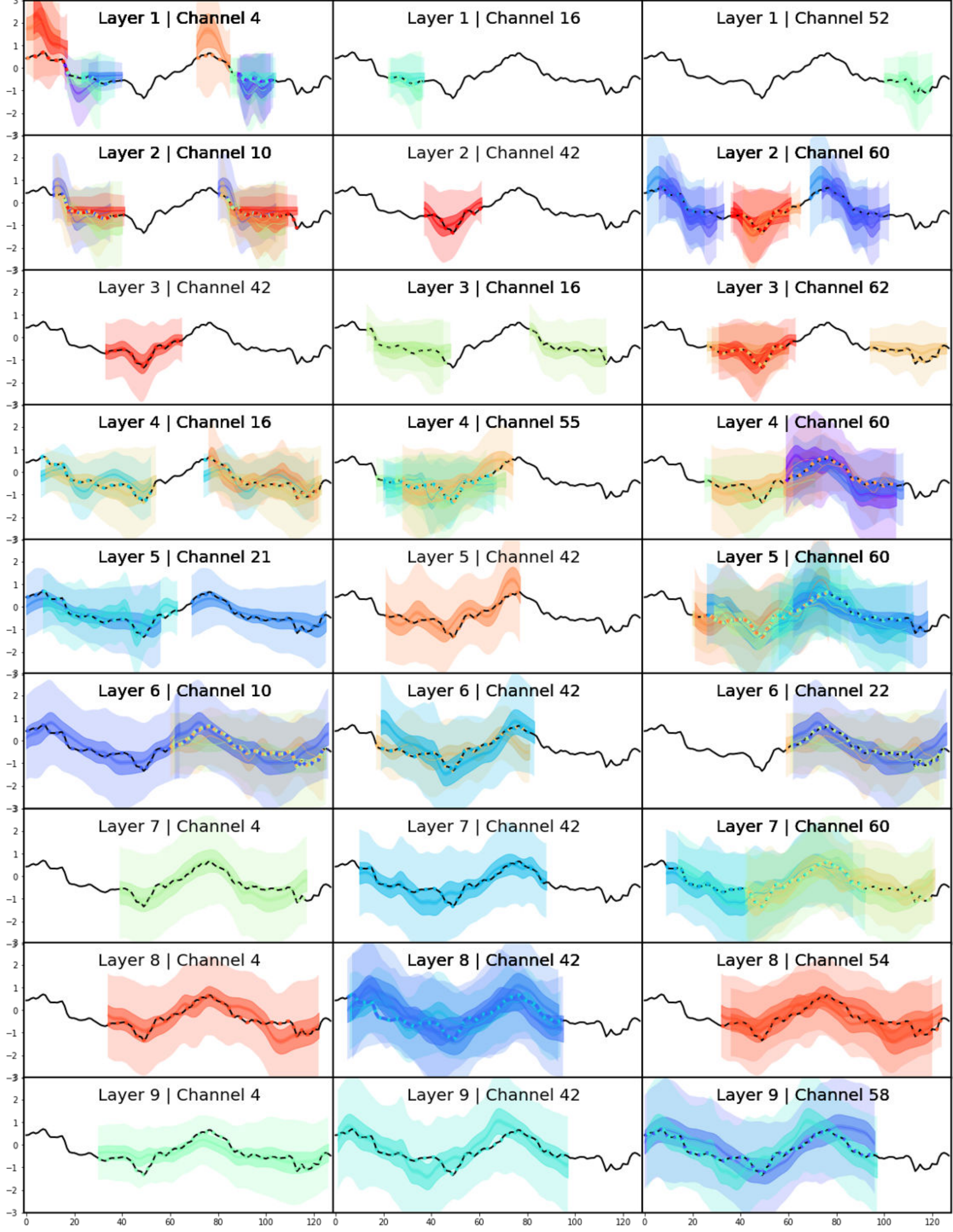}
    \caption{CPHAP result for longer patterns in ResNet Model, HAR Data 7730, Sensor 5}
    \label{fig: resnet_longerfilter_1}
\end{figure}

\begin{table*}[h!]
\caption{ResNet Structure with Shorter Filters}
\label{ResNet Structure with Shorter Filters}
\vskip 0.05in
\begin{center}
\begin{small}
\begin{sc}
\begin{tabular}{lccccccccccc}
\toprule
  &\shortstack{ conv1\\filter}  &\shortstack{ conv2\\filter }  &\shortstack{ conv3\\filter }  &\shortstack{ conv4\\filter }  &\shortstack{ conv5\\filter }  &\shortstack{ conv6\\filter }  &\shortstack{ conv7\\filter }  &\shortstack{ conv8\\filter }  &\shortstack{ conv9\\filter }\\
\midrule
filter size  & 8       & 5       & 3       & 8       & 5       & 3       & 8       & 5       & 3       \\
pattern size & 8       & 12      & 14      & 21      & 25      & 27      & 34      & 38      & 40     \\
\bottomrule
\end{tabular}
\end{sc}
\end{small}
\end{center}
\end{table*}

\begin{figure}[h!]
\vskip -1in
\centering
    \includegraphics[width=0.85\linewidth]{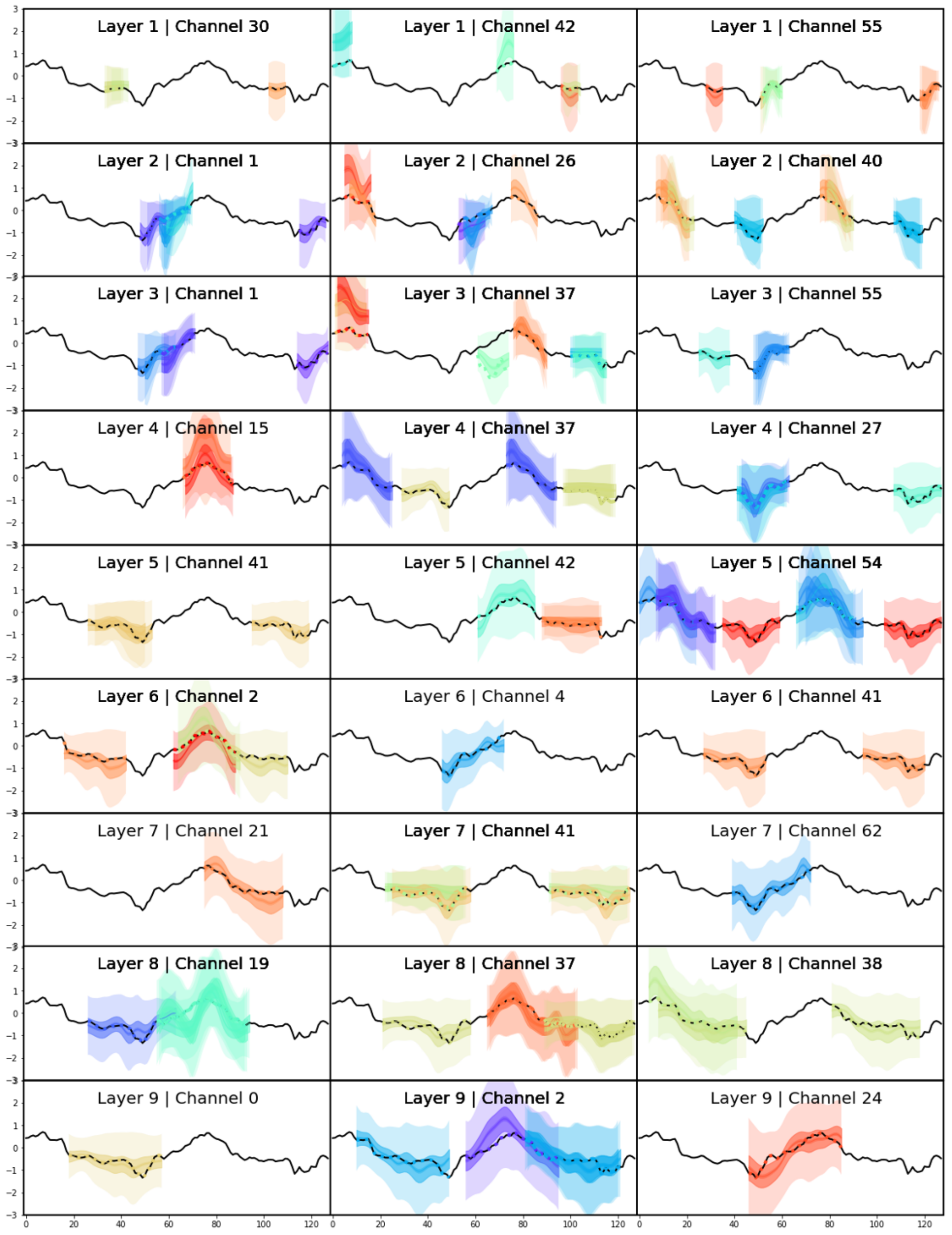}
    \caption{CPHAP result for shorter patterns in ResNet Model, HAR Data 7730, Sensor 5 }
    \label{fig: resnet_har_7730_1}
\end{figure}

\clearpage
\section{CPHAP in Test Dataset}
\hspace{3mm} Figure 9 illustrates how CPHAP works well in the test dataset, which is not used to train the pattern clusters. Figure 9 (1) shows when CPHAP is well matched with new test data. On the other hands, Figure 9 (2) shows examples of less-matched patterns. Note that there are certain points where the actual data deviate from the assigned pattern for each less-matched example. Actually, we can identify that the distribution of each pattern, which is obtained from training dataset, has the large uncertainty on those points in Figure 9 (c). It supports that our framework is valid for capturing uncertainties in sub-sequences that activate specific channels.

\begin{figure}[h]
    \includegraphics[width=\linewidth]{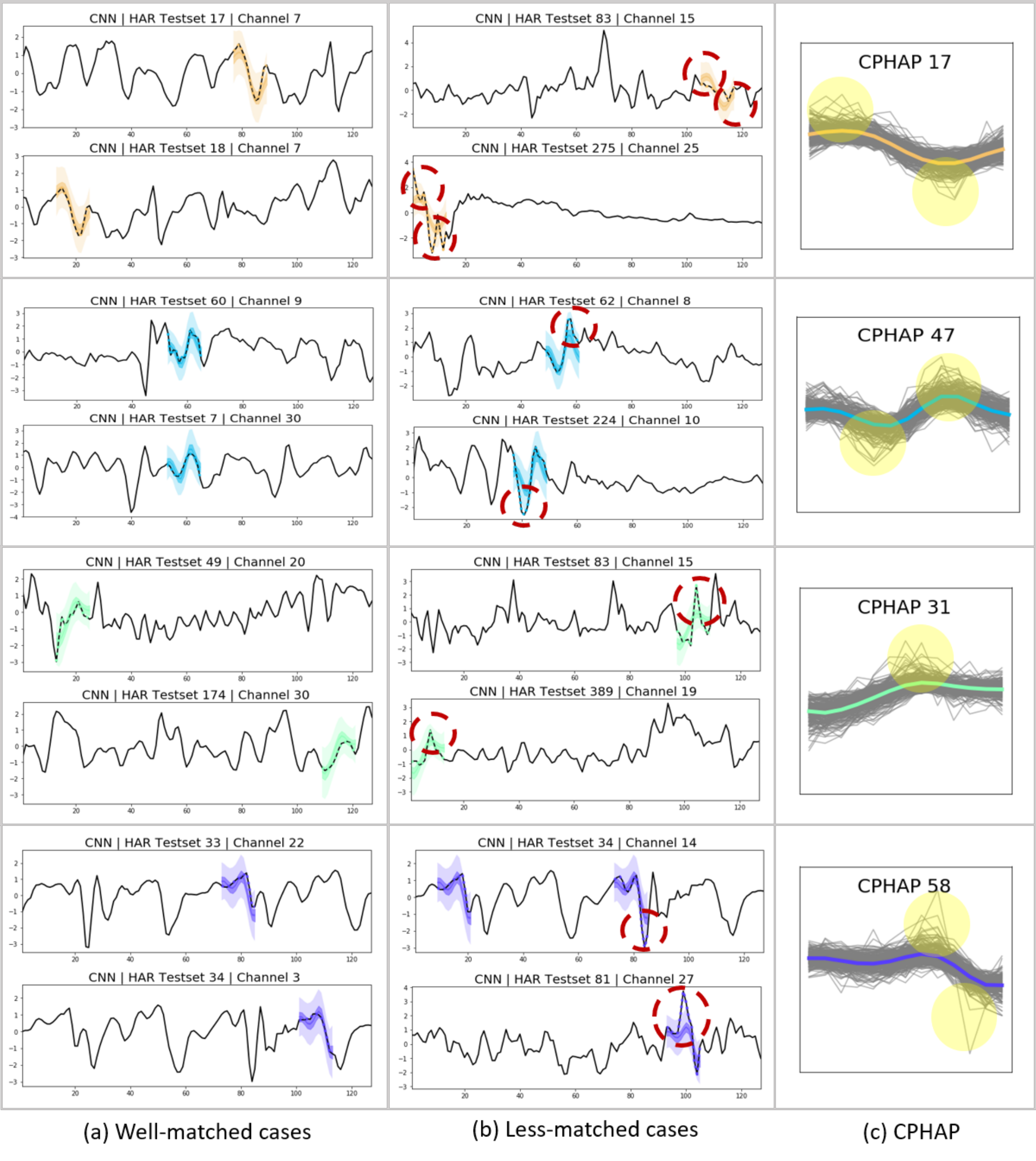}
    \caption{Comparisons between well-matched patterns and less-matched patterns}
    \label{fig: mis_2}
\end{figure}

\clearpage
\section{Patternizing Representative Sub-sequences by Clustering }

\hspace{3mm} We compare several clustering methods to suggest an optimal algorithm for characterizing input sub-sequences with a measure of shape and distribution. We applies various methods, including K-means, Gaussian Mixture Model (GMM), K-shape, and Self Organizing Map (SOM) methods. The results of clustering methods are below.

\hspace{3mm} SOM is an unsupervised learning method by mapping high-dimensional data to a low-dimensional map with keeping the topology of the map space. GMM assumes that there is a certain number of Gaussian distributions, and that each of these distributions represents a cluster. K-means method classifies nearby sequences into identical clusters based on Euclidean distance only. K-shape method is similar to K-means, but specialized for time series: it compares sequences efficiently and computes centroids effectively under scaling and shift invariances.

\hspace{3mm} We choose SOM for two reasons. First, the mean value of each SOM cluster seems to be an valid representation of the sub-sequences belonging to that cluster. Second, the map space of SOM represents spatial meanings. In other words, near cluster groups show similar patterns on the map. We expect this property to help to explain the role of channels in the convolutional neural network.

\hspace{3mm} Figure 10 illustrate the clustering result of SOM, aligning with the map space of SOM. Also, we plot the marginalized clusters along horizontal axis and vertical axes. We can identify that near cluster groups have similar patterns. For the CNN model trained on Uwave dataset, we observe that the clusters detected by channel 1 in layer 2 are cluster 3, 9, 10, 11, 12, 16, 17 and 18 in Figure 10. These clusters are actually close together in the map space.

\begin{figure}[h]
\centering
    \includegraphics[width=0.7\linewidth]{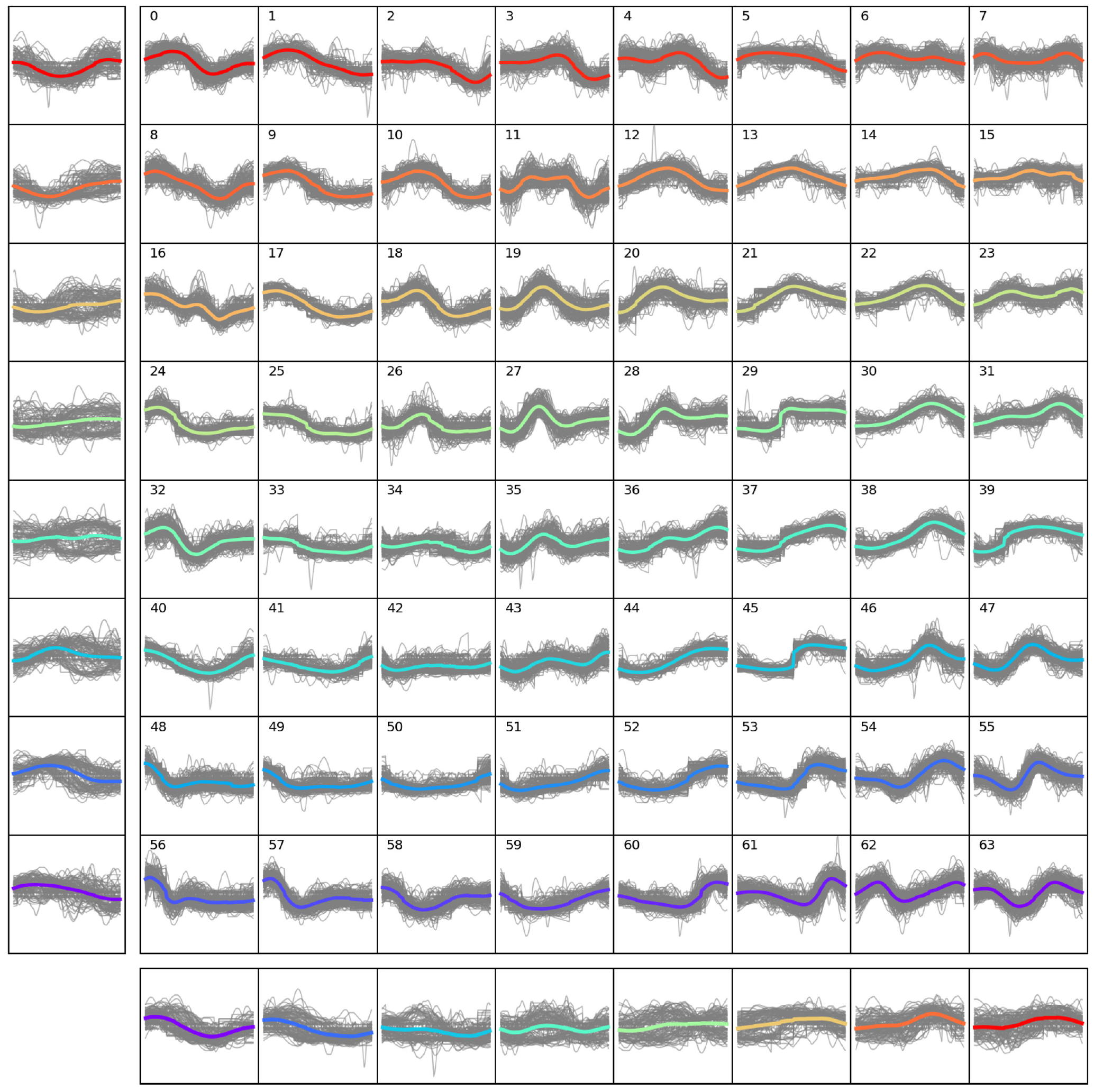}
    \caption{SOM clustering result}
    \label{fig: som_cluster}
\end{figure}

\newpage

\begin{figure}[h!]
\centering
    \includegraphics[width=0.7\linewidth]{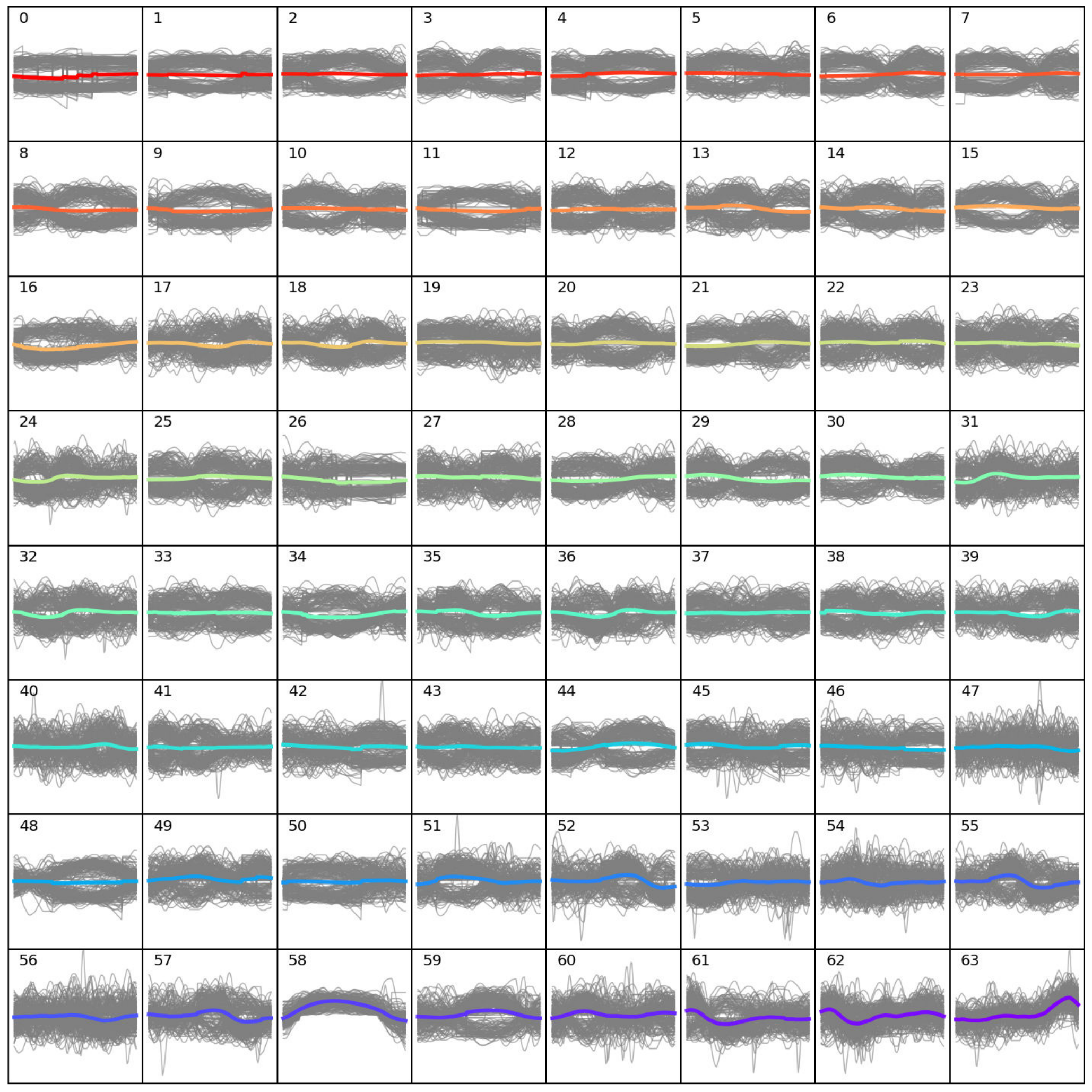}
    \caption{GMM clustering result}
    \label{fig: gmm_cluster}
\end{figure}

\begin{figure}[h!]
\centering
    \includegraphics[width=0.7\linewidth]{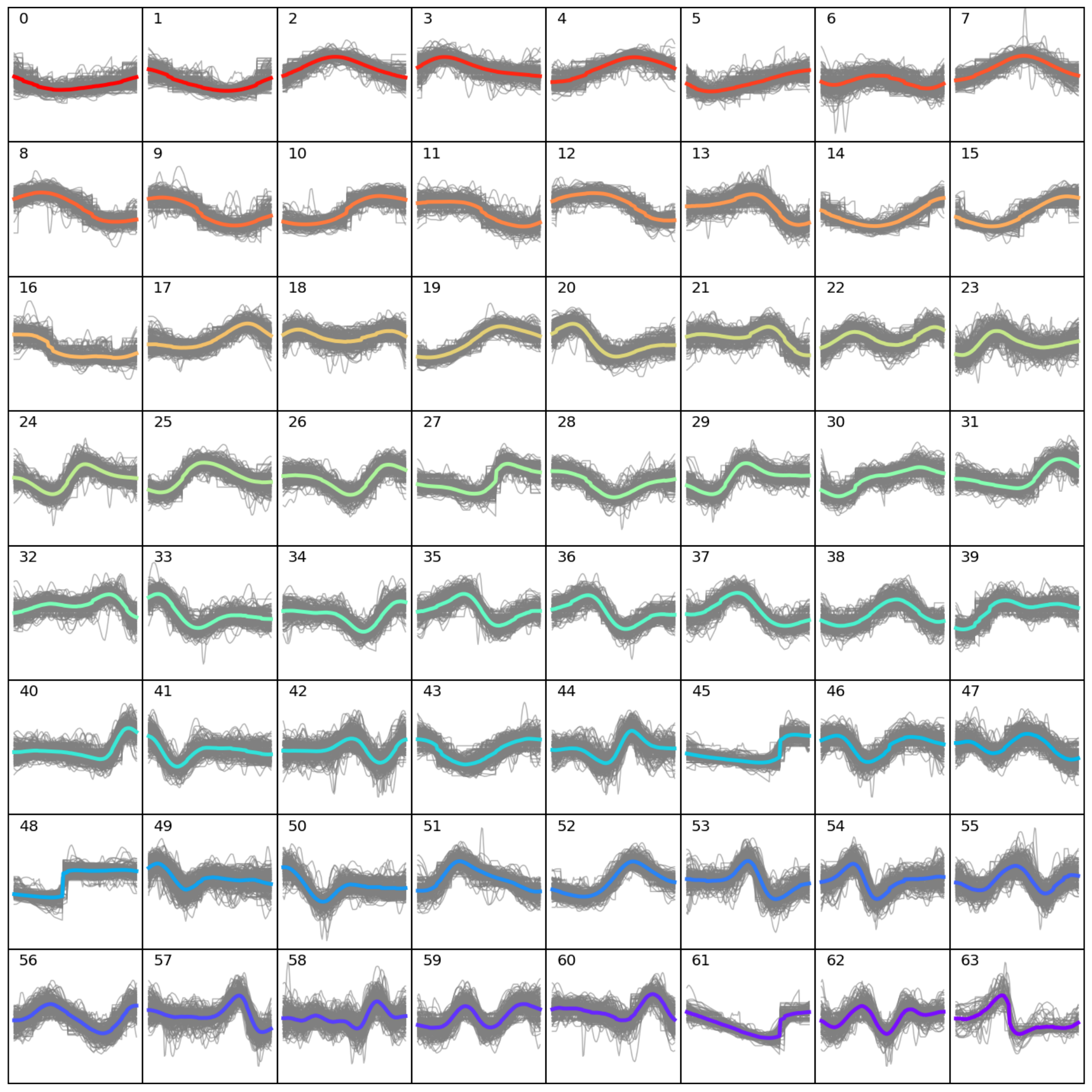}
    \caption{K-means clustering result}
    \label{fig: kmeans_cluster}
\end{figure}

\newpage

\begin{figure}[]
\centering
    \includegraphics[width=0.7\linewidth]{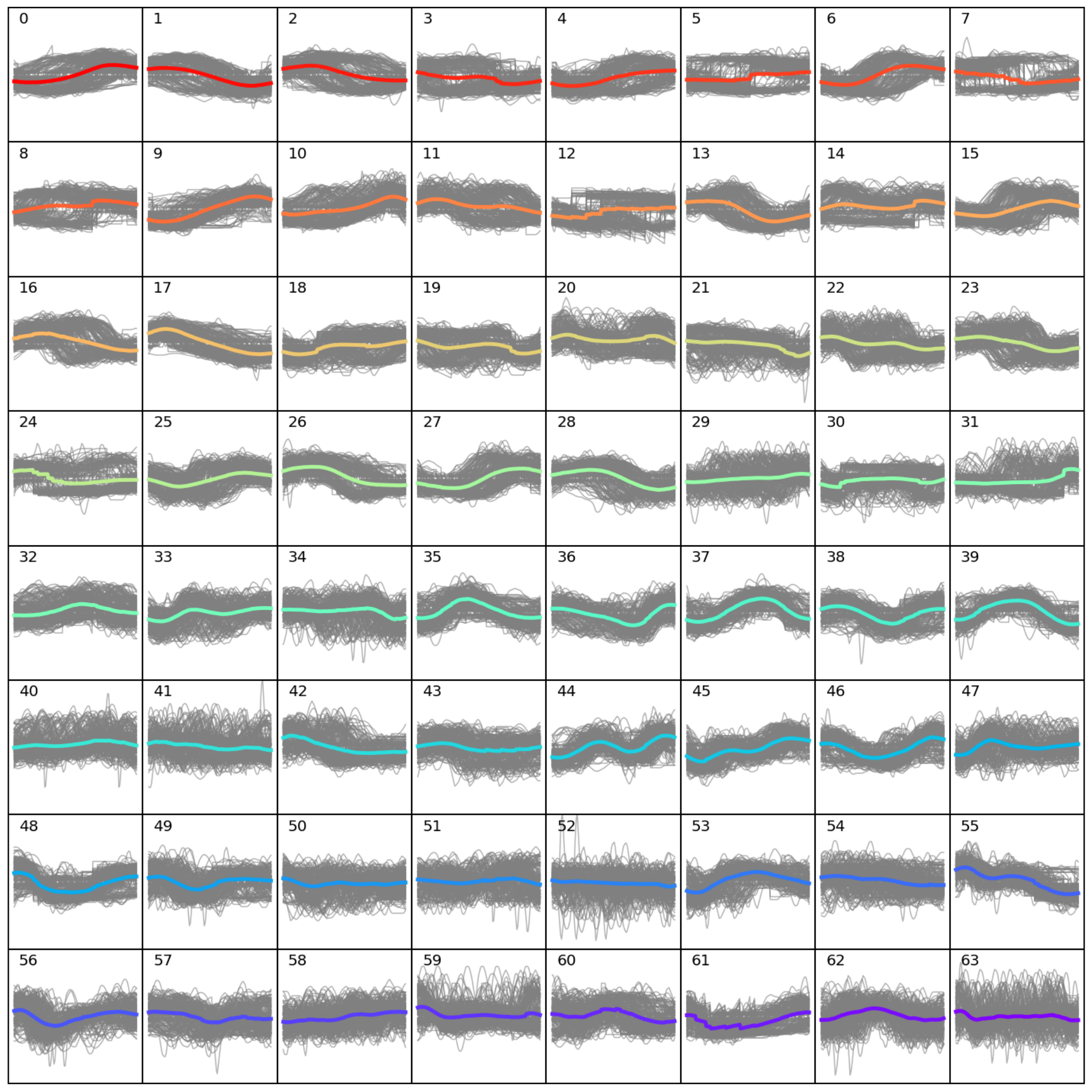}
    \caption{K-shape clustering result}
    \label{fig: kshape_cluster}
    \vspace{15mm}
\end{figure}

\section{Identifying role of the channel in CPHAP}


\begin{figure}[h]
    \includegraphics[width=\linewidth]{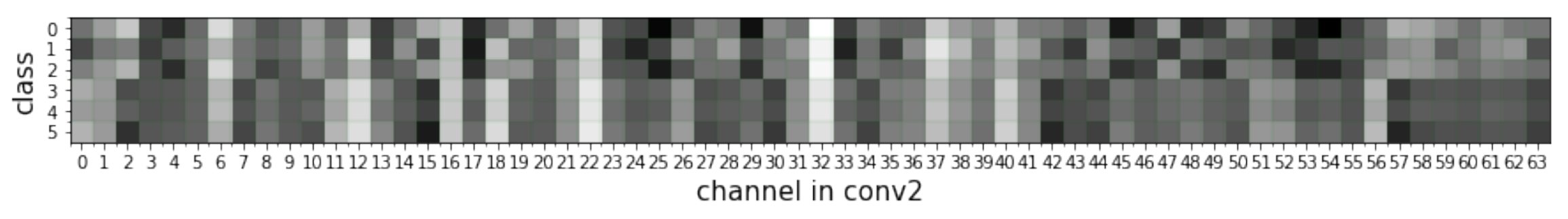}
    \caption{Class-Channel relationship in layer 2 on HAR dataset}
    \label{fig:har_class_pattern}
\end{figure}

\hspace{3mm} Figure 14 is the frequency matrix from Class-channel relationship for HAR dataset. Channel  4, channel 25, channel 29, channel 45, and channel 54 are frequently activated for data in class 0 or class 2. For data in class 5, channel 2, channel 15, channel 42 and channel 57 are activated a lot. Data belonging to class 3 and class 5 seem to have similar class-channel relationship.



\newpage

\section{Perturbation analysis}

\hspace{3mm} We appends more samples for perturbation analysis. The Y-axis in Figure 14 denotes the change in the activation value according to perturbation. That is, the large change of Y value implies that the corresponding algorithm fails to select the appropriate points related to the channel. 

\hspace{3mm} In the case of Gaussian perturbation, preserving the temporal regions selected by CPHAP makes the channel activation more robust for most channels. For some channels, such as channel 15, CPHAP has similar performance to LRP when using Inverse or Zero perturbation. However, CPHAP generally selects more important points than LRP to maintain the activation.

\begin{figure}[h]
\centering
    \includegraphics[width=0.9\linewidth]{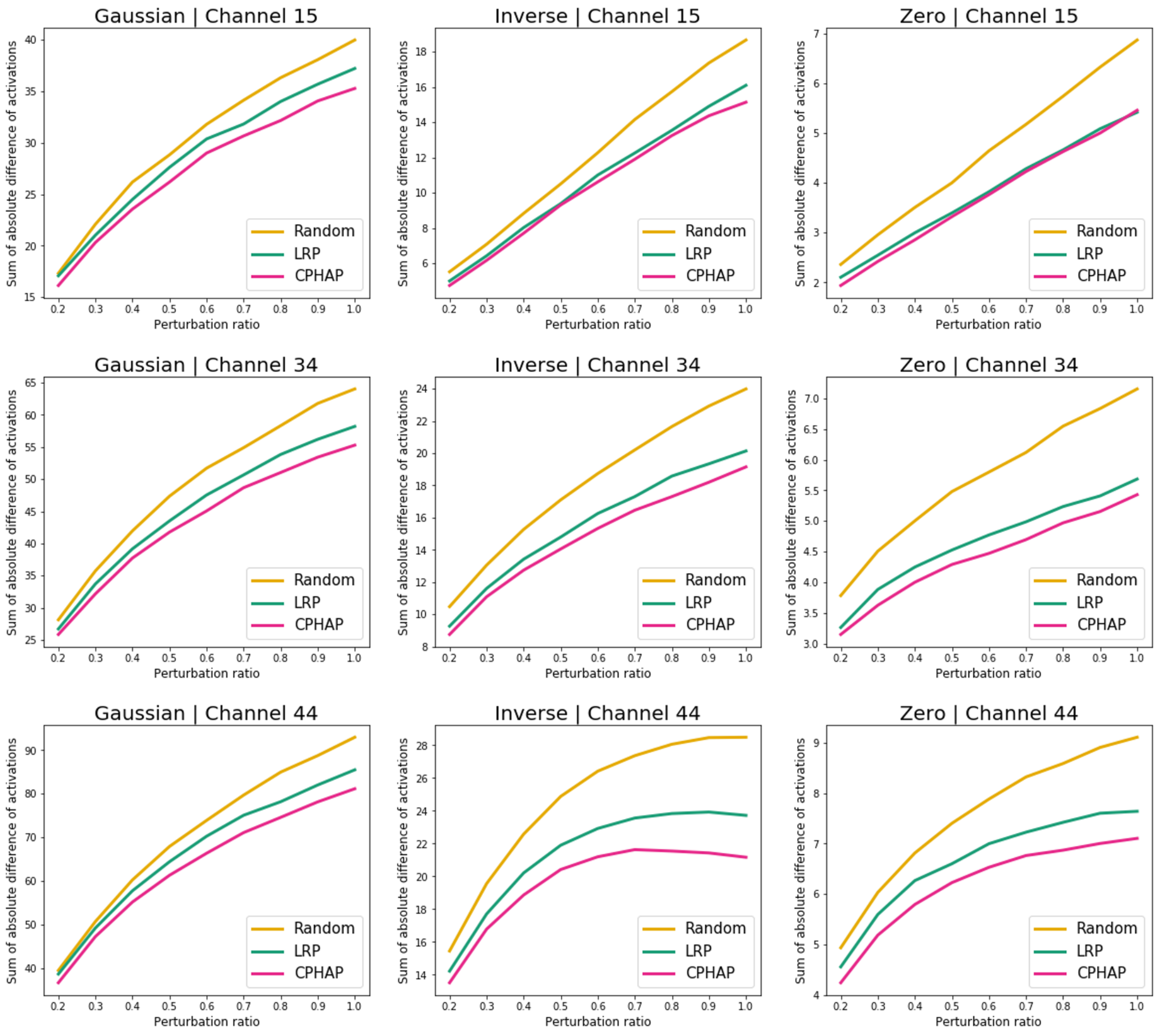}
    \caption{The results of perturbation analysis for sampled channels.}
    \label{fig: perturbation}
\end{figure}

\end{document}